\documentclass[11pt]{article}

\usepackage{hyperref}[citecolor=lightblue]
\hypersetup{
    colorlinks = true,
    citecolor = {magenta},
}
\usepackage[margin=1.0in]{geometry}
\usepackage{microtype}
\usepackage{graphicx}
\expandafter\def\csname ver@subfig.sty\endcsname{}
\usepackage{booktabs} %
\usepackage{float}
\usepackage{amsmath}
\usepackage{amssymb}
\usepackage{mathtools}
\usepackage{amsthm}
\usepackage{mathrsfs}
\usepackage{nicefrac}
\usepackage{dsfont}
\usepackage{enumitem}
\usepackage{graphicx}
\usepackage{float}
\usepackage{hyperref}[citecolor=magenta]
\hypersetup{
    colorlinks = true,
    citecolor = {magenta},
}
\usepackage[utf8]{inputenc} %
\usepackage[T1]{fontenc}    %
\usepackage{hyperref}       %
\usepackage{url}            %
\usepackage{booktabs}       %
\usepackage{amsfonts}       %
\usepackage{nicefrac}       %
\usepackage{microtype}      %
\usepackage{xcolor}         %
\usepackage{textcomp}
\usepackage[toc,page,header]{appendix}
\usepackage{silence}
\usepackage{graphicx}
\usepackage{math_commands}
\usepackage{hyperref}
\usepackage[authoryear]{natbib}

\usepackage{nicefrac}
\usepackage{amsmath}
\usepackage{xfrac}
\usepackage{graphicx}
\usepackage{booktabs}
\usepackage{wrapfig}
\usepackage{units}
\usepackage{amsthm}
\usepackage{enumitem}
\usepackage{bbm}

\usepackage{xspace}

\usepackage{multirow}

\theoremstyle{plain}
\newtheorem{theorem}{Theorem}[section]
\newtheorem{proposition}[theorem]{Proposition}
\newtheorem{lemma}[theorem]{Lemma}
\newtheorem{fact}[theorem]{Fact}
\newtheorem{corollary}[theorem]{Corollary}
\theoremstyle{definition}
\newtheorem{definition}[theorem]{Definition}

\theoremstyle{remark}

\usepackage{tikz}
\usetikzlibrary{shapes}
\usetikzlibrary{decorations.pathreplacing,shapes}
\newcommand*\circled[1]{\tikz[baseline=(char.base)]{
            \node[shape=circle,draw,inner sep=2pt] (char) {#1};}}

\let\originalleft\left
\let\originalright\right
\renewcommand{\left}{\mathopen{}\mathclose\bgroup\originalleft}
\renewcommand{\right}{\aftergroup\egroup\originalright}

\renewcommand{\epsilon}{\varepsilon}

\title{{Lower Bounds for Public-Private Learning under Distribution Shift}}

\begin{document}

\author{
	Amrith Setlur\thanks{Carnegie Mellon University. Supported by JP Morgan AI PhD Fellowship.  \texttt{asetlur@andrew.cmu.edu}}
 	\and
        Pratiksha Thaker\thanks{Carnegie Mellon University. Supported by Carnegie Bosch Institute Fellowship.  \texttt{pthaker@andrew.cmu.edu}}
	\and
	Jonathan Ullman\thanks{Khoury College of Computer Sciences, Northeastern University.  Supported by NSF awards CNS-232692 and CNS-2247484. \texttt{jullman@ccs.neu.edu}}
}
\date{}
\maketitle 

\begin{abstract}
The most effective differentially private machine learning algorithms in practice rely on an additional source of purportedly public data.  This paradigm is most interesting when the two sources combine to be more than the sum of their parts.  However, there are settings such as mean estimation where we have strong lower bounds, showing that when the two data sources have the same distribution, there is no complementary value to combining the two data sources.  In this work we extend the known lower bounds for public-private learning to setting where the two data sources exhibit significant distribution shift.  Our results apply to both Gaussian mean estimation where the two distributions have different means, and to Gaussian linear regression where the two distributions exhibit parameter shift. We find that when the shift is small (relative to the desired accuracy), either public or private data must be sufficiently abundant to estimate the private parameter. Conversely, when the shift is large, public data provides no benefit.
\end{abstract}

\section{Introduction}
\label{sec:introduction}
Differential privacy~\citep{dwork2006calibrating} is a formal privacy framework that makes it possible to perform statistical analysis or train large models while giving a strong guarantee of privacy to the individuals who contribute data.  Informally, a differentially private algorithm does not reveal too much information about any one individual's data.  While differential privacy has been deployed successfully for statistics and machine learning, many of the most effective deployments work in a \emph{public-private} setting in which there is access to an additional source of \emph{public} data that can be used along with the private data.  In this setting, the overall algorithm need not be fully differentially private---its output can reveal arbitrary information about the public data, but must limit the amount of information revealed about the private data.  For example, the most effective differentially private predictive and generative models are often initialized with a large pretrained model that is not differentially private~\citep{yu2021differentially,li2021large}, and the best models trained from scratch with only private data have much worse accuracy.  This paradigm is most interesting when the public and private data combine to be more than the sum of their parts---by combining public and private data we can train an accurate model even though neither source on its own is large or high-quality enough to train an accurate model.

However, not every application provides us with a useful source of public data, and there are also important questions about whether these purportedly public datasets raise privacy concerns of their own~\citep{tramer2022considerations}.  As a result there is a growing body of theoretical work trying to understand whether the use of public data is \emph{necessary} for private training, or can be replaced with better algorithms.  These results give inconclusive answers, either showing that combining private and public data \emph{does help} (e.g.\ \citep{bie2022private,ganesh2023public}) or \emph{doesn't help} (e.g.\ \citep{bassily2020private,ullah2024public,lowyoptimal}) depending on the specifics of the learning problem.

The prior work that most directly relates to ours studies \emph{mean estimation} as a test case.  Here we are given $n$ \emph{private samples} and $m$ \emph{public samples}, all drawn i.i.d.\ from a distribution $P$, and our goal is to design an algorithm that provides DP for the private samples but can do anything it wants with the public samples and estimates $\mu = \mathbb{E}[P]$.  In this setting, it is known that combining public and private data doesn't help make more effective use of the private data~\citep{bassily2020private}.  Throughout this work, when we say that combining private and public data \emph{doesn't help}, we mean roughly that whenever we are able to solve the learning problem using the combination of public and private data, then we could also have solved the learning problem (perhaps up to some small contant loss in accuracy) with just the public data or the private data alone.

\subsection{Contributions.}
The above results raise two questions that we address in this work.  First, prior work considers public and private data from the \emph{same distribution}; we consider settings where the public and private data exhibit \emph{distribution shift}, which is a more realistic setting for practical applications~\citep{tramer2022considerations}.  Second, we consider whether this phenomenon---that public data doesn't help make more effective use of private data---extends beyond mean estimation to more realistic supervised learning problems like \emph{linear regression}.

\medskip\noindent{\em \bfseries Mean Estimation with Distribution Shift.} In this setup, we assume the private samples are drawn from a distribution $P$ and the public samples are drawn from some related distribution $Q \neq P$, and our goal is to estimate the mean of $P$.  We know from prior work that when $Q$ and $P$ are the same, then there is no value to data from $Q$ unless we already have enough samples from $Q$ to solve the problem, and giving more samples from $Q$ can only improve the achievable accuracy.  We also know that holding the number of samples from $Q$ fixed, shifting $Q$ farther from $P$ can only make the achievable accuracy worse.  However, a priori it might be the case that shifting $Q$ away from $P$, but giving many samples from $Q$, would allow us to find a sweet spot where the samples from $Q$ are not enough on their own because of the shift, and samples from $P$ are not enough because of privacy, but the two can be combined to achieve the desired accuracy.

Our first main result shows that this is not the case---that public data still doesn't help, even in settings where the public and private data come from different, but related, distributions. That is, we show that if we are able to estimate the mean up to some specified error, then \emph{either} (1) the public data is abundant enough and its distribution $Q$ is close enough to $P$ that we can estimate the mean using only public data, \emph{or} (2) the private data is abundant enough that we can privately estimate the mean using only the private data.

 Specifically, we consider a setting where $P$ and $Q$ are both Gaussians with $P = N(\mu_{P},\mathbf{I}_d)$ and $Q = N(\mu_{Q},\mathbf{I}_d)$ we promise that $\| \mu_{P} - \mu_{Q} \|_2 \leq \tau$, and our goal is to find an estimate $\hat\mu_{P}$ such that $\| \hat\mu_{P} - \mu_{P} \|_2 \leq \alpha$.  We assume access to $n$ private samples from $P$ and $m$ public samples from $Q$.  Note that if we have access to only private samples from $P$, then we could solve the problem if and only if $n = \Omega(\nicefrac{d}{\alpha \eps})$.  Also note that if $\tau \leq \alpha$ then, up to a factor of $2$ in the error, we can simply use the public data as if it were from the same distribution as the public data.  Thus, if we had access to only public samples from $Q$ then we would need that both $\tau < \alpha$ and $m = \Omega(\nicefrac{d}{\alpha^2})$.  Our result shows that, up to constant factors, we need either the private or public data to be sufficient on their own, and there is no complementarity between them.
\vfill\newpage
\begin{theorem}[Informal; mean estimation when $P\neq Q$]
    \label{thm:lb-gaussian-dist-shift-informal}
    \phantom{.}
    \begin{enumerate}
        \item When the distribution shift is small, (i.e., $\tau \lesssim \alpha$) then to solve the mean estimation problem with error $\alpha$ in $\ell_2$, we must have either enough public data alone to estimate the mean (i.e., $m = \Omega(\nicefrac{d}{\alpha^2})$), or enough total data to solve the mean estimation problem with differential privacy i.e., $n+m = \Omega(\nicefrac{d}{\alpha\varepsilon} + \nicefrac{d}{\alpha^2})$.  
        
        \item When the distribution shift is large, (i.e., $\tau \gtrsim \alpha$), then the entire sample complexity burden falls on private data and we must have enough private data to solve the mean estimation problem with differential privacy (i.e., $n = \Omega(\nicefrac{d}{\alpha\varepsilon} + \nicefrac{d}{\alpha^2})$.
    \end{enumerate}
\end{theorem}

We detail this theorem and sketch its proof in Section~\ref{sec:gaussian-dist-shift}.

\medskip\noindent{\em \bfseries Linear Regression without Distribution Shift.} Here, we consider the richer setting of private linear regression with Gaussian covariates and Gaussian errors.  That is, the data consist of pairs $(\vec{x},y)$ where the features $x$ are drawn from a standard multivariate Gaussian $N(0,
\mathbf{I}_{d})$ and $y|\vec{x} =  \langle \beta, \vec{x} \rangle + \nu$ for some ground truth predictor $\beta$ and error $\nu$ drawn from a standard Gaussian $N(0,1)$.  Our goal is to estimate the optimal predictor $\beta$ with small excess squared prediction error, which in this setting is equivalent to minimizing $\|\hat\beta - \beta\|_2^2$.  For now we assume that the public and private data are both sampled from the same distribution.

Our second main result shows that the same qualitative phenomenon holds in this setting, that either the public data is sufficient to train a good predictor on its own, or the private data is sufficient to privately train a good predictor on its own.

\begin{theorem}[Informal; linear regression when $P = Q$] 
    \label{thm:lb-linreg-no-shift-informal}
    To solve the linear regression problem with excess squared prediction error $\alpha^2$, we either must have enough public samples alone to solve the linear regression problem (i.e., $m = \Omega(\nicefrac{d}{\alpha^2})$), or we must have enough total data to solve the problem with differential privacy (i.e., $n+m = \Omega(\nicefrac{d}{\alpha\eps} + \nicefrac{d}{\alpha^2})$).
\end{theorem}

\medskip\noindent{\em \bfseries Linear Regression with Distribution Shift.} Finally, we consider Gaussian linear regression with \emph{parameter shift} between the public and private distributions.  The only change to the setting above is that, for the private data, we have $y|\vec{x} =  \langle \beta_{P}, \vec{x} \rangle + \nu$, and for the public data, we have $y|\vec{x} =  \langle \beta_{Q}, \vec{x} \rangle + \nu$ for $\beta_{P} \neq \beta_{Q}$.  In analogy to the case of mean estimation, we will parameterize the degree of distribution shift by a parameter $\tau$ with the promise that $\| \beta_{P} - \beta_{Q} \|_2^2 \leq \tau^2$. Note that, analogous to the case of mean estimation, if $\tau \leq \alpha$, then up to a factor of $2$ in the error, we can treat the public data as if it came from the same distribution as the private data.  The goal is still to estimate $\beta_{P}$ with small excess squared prediction error, or equivalently to minimize $\|\hat\beta - \beta_{P}\|_2^2$.

Analogous to the case of mean estimation with distribution shift and linear regression without distribution shift, our third main result shows that combining public and private data doesn't help in this setting.

\begin{theorem}[Informal; linear regression when $P \neq Q$] 
    \label{thm:lb-linreg-dist-shift-informal}
    \phantom{.}
    \begin{enumerate}
        \item When the distribution shift is small, (i.e., $\tau \lesssim \alpha$) then to solve linear regression with excess squared prediction error $\alpha^2$, we must have either enough public to solve the linear regression problem (i.e., $m = \Omega(\nicefrac{d}{\alpha^2})$), or enough total data to solve the mean estimation problem with differential privacy (i.e., $n+m = \Omega(\nicefrac{d}{\alpha\varepsilon} + \nicefrac{d}{\alpha^2})$).  
        
        \item When the distribution shift is large, (i.e., $\tau \gtrsim \alpha$), then the entire sample complexity burden falls on private data and we must have enough private data to solve the mean estimation problem with differential privacy (i.e., $n = \Omega(\nicefrac{d}{\alpha\varepsilon} + \nicefrac{d}{\alpha^2})$.
    \end{enumerate}
\end{theorem}

\medskip\noindent{\em \bfseries Proof Techniques.} Our work builds on the fingerprinting method~\citep{ullman2013answering,bun2014fingerprinting,dwork2015robust}, which has been widely applied to proving lower bounds in differential, including in the public-private setting~\citep{bassily2020private,ullah2024public,lowyoptimal}.  In particular, we adopt the recently introduced \emph{Bayesian perspective on the fingerprinting method}~\cite{narayanan2023better}.  As a starting point for our work, we can show that the Bayesian perspective gives an especially clean and simple way to establish lower bounds in the most basic setting of mean estimation with public-private data \emph{without distribution shift,} recovering the result from~\citep{bassily2020private}.  This new proof allows us to extend the results for mean estimation to settings with distribution shift (Theorem~\ref{thm:lb-gaussian-dist-shift-informal}) by giving an easy way to choose a hard distribution for the public data and easily incorporating the influence of the public samples on the distribution of the private data using conjugate priors.  This viewpoint also allows us to extend the lower bounds to settings like linear regression (Theorem~\ref{thm:lb-linreg-no-shift-informal}), where lower bounds were previously only known for the purely private setting~\citep{cai2021cost} and not the public-private setting.  Finally, we can combine these two extensions can be combined to the setting of linear regression with parameter shift (Theorem~\ref{thm:lb-linreg-dist-shift-informal}).  A key technique we employ in this setting is to reinterpret the parameter shift in linear regression as a form of non i.i.d.\ noise on the labels and then employ methods from generalized least-squares estimation.  See Section~\ref{sec:proof-overview} for a more thorough discussion of these techniques.

\subsection{Related Work}

The most closely related works to ours are \citep{bassily2020private,ullah2024public,lowyoptimal}, which give lower bounds for mean estimation and stochastic convex optimization in the public-private setting using fingerprinting techniques, but only consider the setting where the public and private data come from the same distribution.  We now review the related work on public-private estimation and machine learning as well as on fingerprinting lower bounds in differential privacy.

\smallskip{\em \bfseries Public-private statistical estimation.}
A number of prior works have theoretically studied the benefits of public data in other settings, including mean estimation~\citep{avent2020power}, query release~\citep{liu2021leveraging, bassily2020private, fuentes2024joint}, convex optimization when gradients lie in a low-rank subspace~\citep{kairouz2021nearly, amid2022public}, and non-convex optimization~\citep{ganesh2023public}.  In particular~\citep{ganesh2023public} gives a non-convex optimization setting where public data is necessary to achieve high accuracy.  Typically these positive results for combining public and private data arise in settings where the public data can be used for some kind of dimensionality reduction or pretraining, where the dimensionality reduction step is often much more costly under privacy constraints than when using public data.  In contrast to our work, many of these positive results formally assume that the public and private data come from the same distribution and doesn't directly address the question of how close the public and private data need to be in order for these positive results to hold.

The combination of public and private data has also been used effectively for practical machine learning in a number of domains~\citep{abadi2016deep, papernot2019making, tramer2020differentially, luo2021scalable, yu2021differentially, de2022unlocking, ke2024convergence, ganesh2023public, kurakin2022toward, mehta2022large, yu2021large, li2021large, yu2021differentially, arora2022can, golatkar2022mixed, kurakin2022toward, he2022exploring, bu2023differentially, ginart2022submix}, however these works are primarily empirical, whereas our work is focused on understanding whether combining public and private data is necessary.

\smallskip{\em \bfseries Fingerprinting lower bounds in differential privacy.} Fingerprinting methods have become the standard tool for establishing lower bounds on the cost of differentially private statistical estimation, particularly for high-dimensional problems.  This method was first introduced in differential privacy in \citep{ullman2013answering, bun2014fingerprinting}.  This method was subsequently refined and used to prove lower bounds for a wide variety of problems~\citep{dwork2014analyze, bassily2014private, dwork2015robust, steinke2015between, steinke2017tight, kamath2019privately, cai2021cost, narayanan2022tight, kamath2022new, cai2023score, narayanan2023better, peter2024smooth, aliakbarpour2024privacy, portella2024lower}. Relevant for our work is the the so-called score attack for parameter estimation~\citep{cai2023score}, which we adopt for our linear regression proofs.  This general method has also been used for proving lower bounds outside of privacy, specifically for problems in adaptive data analysis~\citep{hardt2014preventing, steinke2015interactive, ullman2018limits, lyu2024fingerprinting} and the information-cost of learning~\citep{attias2024information}.  Fingerprinting methods are also closely related to membership-inference attacks~\citep{dwork2015robust} and privacy auditing~\citep{jagielski2020auditing}.

\section{Technical Overview}
\label{sec:proof-overview}

In this section, we describe the structure of our proofs at a high level and highlight the new elements we introduce.  We present a general proof technique for proving lower bounds for public-private computations under distribution shifts.  Our approach starts with the Bayesian approach for proving lower bounds in the private-only setting from~\citet{narayanan2023better}.  We show that this approach is particularly well suited to proving lower bounds under distribution shift between public and private data.  To start, we will review this Bayesian approach specific to the setting of Gaussian mean estimation, and then we will discuss how to incorporate distribution shift to prove our new results.  Later we will discuss the key ideas in adapting this proof to the linear regression setting.

\medskip\noindent
{\em \bfseries Problem Setup and Prior Work: Private-Only Mean Estimation.} 
In the private mean estimation problem we have a dataset $X$ consisting of $n$ samples $X_1,\dots,X_n$ drawn i.i.d.\ from a spherical multivariate normal distribution in $P = N(\mu, \mathbf{I}_d)$.  Our goal is to design an $(\eps,\delta)$-differentially private algorithm $M(X)$ that returns an estimate $\hat\mu$ such that $\|\hat\mu - \mu\|_2 \leq \alpha$.  Without privacy, the standard algorithm is to compute the empirical mean $\hat\mu = \bar{X} = \frac{1}{n} \sum_i X_i$ and we will satisfy the accuracy goal as long as $n = \Omega(\nicefrac{d}{\alpha^2})$, which is also optimal.  We want to know how the sample requirement changes when we require that the estimator is differentially private.

The main technique for proving such lower bounds is the so-called \emph{fingerprinting method}.  In this method we design a \emph{test statistic} $Z$ defined as
\begin{align}
Z_i = \langle X_i - \mu, \hat\mu - \mu\rangle \textrm{ and } Z = \frac{1}{n} \sum_i Z_i = \langle \bar{X} - \mu, \hat\mu - \mu\rangle \label{eq:test-statistic-zi}
\end{align}
Intuitively each $Z_i$ measures the covariance between the estimate $\hat\mu$ and a single data point $X_i$.  The reason this test statistic is useful is that: (1) an accurate $\hat\mu$ should depend on many of the data points $X_i$, so we would expect the sum of the correlations $Z$ to be large on average, and (2) a private $\hat\mu$ should not depend significantly on the data points $X_i$, so we would expect sum of the correlations $Z$ to be small on average.  Using these two observations, we can try to derive a contribution showing that accurate and private estimation is impossible for certain choices of parameters.  

Formalizing and proving that the statistic $Z$ is \emph{small} due to privacy is relatively straightforward using the definition of differential privacy.  The tricky part is formalizing and proving that the statistic $Z$ is large for any mechanism that is sufficiently accurate, and this is the crux of the argument, sometimes called the \emph{fingerprinting lemma}.  Although there are many proofs, we outline the main ideas in the proof of ~\citet{narayanan2023better} as a starting point, which starts by writing the  Markov chain
\begin{equation*}
\mu \longrightarrow X_1,\dots,X_n \longrightarrow \hat\mu
\end{equation*}
The key property of this Markov chain is that $\mu$ and $\hat\mu$ are independent conditioned on the data $X$.  Now, a series of calculations gives us a lower bound on $\mathbb{E}_{\mu,X,\hat\mu}[Z]$ that will depend on specific properties of: (1) the \emph{posterior mean} $\mathbb{E}[\mu \mid X]$, (2) the \emph{posterior variance} $\mathit{Var}[\mu \mid X]$, and (3) the accuracy of the estimate $\|\hat\mu - \mu\|_2$.  The third quantity is something we control by assumption and the first two quantities can be controlled by understanding the posterior distribution of $\mu$ conditioned on $X$.  This framework makes it clear that we should choose a \emph{conjugate prior} so that the posterior $\mu \mid X$ has a nice closed form.  In this example, the appropriate conjugate prior is itself a Gaussian so that $\mu \sim N(0,\sigma^2 \mathbf{I}_d)$ for an appropriate choice of $\sigma$, so that $\mu \mid X$ is also Gaussian with some mean and variance we can write down, although the specific calculations are beyond the scope of this overview.
\\

\noindent{\em \bfseries Incorporating Public Data and Distribution Shift (Thm~\ref{thm:lb-gaussian-dist-shift-informal}).}
Now that we have the basic framework in hand, we can discuss why this approach is ideal for incorporating public data into the lower bounds.  Now we have the original distribution $P$ with mean $\mu$ and samples $X_1,\dots,X_n$ that are private.  We also have additional public data $X_{n+1},\dots,X_{n+m}$ samples i.i.d.\ from the same distribution, so the total dataset size is $N = n+m$.  We want to design a mechanism $M(X_1,\dots,X_n,X_{n+1},\dots,X_{n+m})$ that is differentially private as a function of the first $n$ samples.  

Although our formal proof diverges slightly from this outline, for this overview, we can think about the same test statistic $Z = \sum_{i=1}^{n+m} Z_i$ computed over both the public and private samples and look at how the proof does and doesn't change.  First, when we try to establish an upper bound on $Z$, we had used the fact that the mechanism is differentially private as a function of each of its samples, which is no longer true.  However, we can separately bound $\sum_{i=1}^{n} Z_i$ using privacy and get a weaker bound on $\sum_{i=n+1}^{n+m} Z_i$ just using the variance of the distribution.  When we try to establish a lower bound on $Z$, we didn't rely on privacy so not too much changes except that we now have $n+m$ samples instead of just $n$ samples, which slightly affects the posterior mean and covariance.  Although the specific calculations are beyond the scope of this overview, we can use this outline to give a simpler proof of the lower bounds in~\citep{bassily2020private,ullah2024public,lowyoptimal} (captured as the special case of Theorem~\ref{thm:lb-gaussian-dist-shift-informal} where $\tau = 0$).

Now we turn to incorporating distribution shift.  Here, we now assume that the private samples $X_1,\dots,X_{n}$ are drawn from $P = N(\mu_{P},\mathbf{I}_d)$ and the public samples $X_{n+1},\dots,X_{n+m}$ are drawn from $Q = N(\mu_{Q},\mathbf{I}_d)$ where $\| \mu_{P} - \mu_{Q} \|_2 \leq \tau$.  Note that increasing $\tau$ makes the public data less useful and thus should make the problem harder, and thus our goal is ultimately to prove stronger lower bounds on the number of private samples we need to solve the problem.

Once again we consider the same test statistic $Z = \sum_{i=1}^{n+m} Z_i$ and look at how the proof does and doesn't change compared to the above.  In this case the proof of the upper bound on $Z$ is conceptually unchanged compared to the case with no distribution shift.  The main conceptual difference is in the proof of the lower bound on $Z$ and where we require a novel idea.  First, we think of $\mu_{Q} = \mu_{P} + v$ for some random variable $v$ such that $\|v\|_2 \leq \tau$ with high probability, and we will choose the distribution of $v$ later.  Now, if we elide the variable $v$, we have the Markov chain
\begin{equation*}
\mu_{P} \longrightarrow X_1,\dots,X_n,X_{n+1},\dots,X_{n+m} \longrightarrow \hat\mu
\end{equation*}
where all these random variables are fully specified and where $X$ contains all the public and private data.  As before, we want to understand the test statistic $Z$ by using properties of the mean and variance of the posterior distribution $\mu_{P} \mid X$.  Thus, the key idea is to choose the random variable $v$ determining the distribution shift in such a way that $\mu_{P}$ has a nice distribution.  Specifically, what we do is sample $v$ according to a normal $N(0,\nicefrac{\tau^2}{d} \cdot \mathbf{I}_{d})$ so that $\|v\|_2 \leq \tau$ with high probability and so that the distribution
\begin{equation*}
(\mu_{P},X_1,\dots,X_{n},X_{n+1},\dots,X_{n+m})
\end{equation*}
becomes jointly normal in $d(n+m+1)$ variables with a particular block diagonal structure that makes it easy to work with.  As a result the conditional distribution $\mu_{P} \mid X$ is now normal with a particular mean and covariance that we can write down and use to complete the analysis, although the exact calculations are again beyond the scope of this overview.

\medskip\noindent
{\em \bfseries Extending to Linear Regression (Thms~\ref{thm:lb-linreg-no-shift-informal} and~\ref{thm:lb-linreg-dist-shift-informal}).}
An advantage of our proof structure is that is makes it easy to extend our lower bounds to richer problems like Gaussian linear regression.  Here we follow roughly the same outline as mean estimation: (1) we use a suitable test statistic to prove a lower bound for private-only linear regression, in this case coming from~\citep{cai2021cost}, (2) we derive a simplified Bayesian version of the main fingerprinting lemma for that setting that depends on the distribution of the true parameter conditioned on the data and use a suitable conjugate prior to make this choice easy to work with, (3) we extend this fingerprinting lemma to the setting with additional public data, (4) we extend this fingerprinting lemma to the setting with distribution shift by choosing a random shift from a suitable conjugate prior so that the posterior of the true parameter conditioned on both the public and private data has a closed form that is simple enough to analyze.

The main difference between the two settings is in the choice of test statistic and in carefully defining which parts of the experiment we fix and which we don't.  First, in the private-only setting we have a distribution $P$ over labeled samples $(\vec{x},y)$ where we assume $\vec{x}$ has distribution $N(0,\mathbf{I}_d)$ and $y|\vec{x} = \langle \vec{x},\beta \rangle + \nu$ where the error $\nu$ has distribution $N(0,1)$.  Here $\beta$ is the true parameter we want to estimate via a differentially private algorithm $\hat\beta$.

To prove a tight lower bound for this setting,~\citep{cai2021cost} introduced a suitable test statistic, although the specifics are not important for this discussion.  We start by giving a simplified Bayesian version of their analysis.  One key step in our analysis is to think of the feature vectors $\vec{x}_i$ as \emph{fixed} and consider the Markov chain
\begin{equation*}
\beta \longrightarrow y_1,\dots,y_n \longrightarrow \hat\beta
\end{equation*}
Note that we will have to average over the vectors $\vec{x}_i$ at some point, but since they are normally distributed, the covariance of these vectors is tightly concentrated around $\mathbf{I}_d$, so we don't think of this as an important source of randomness.  The advantage of considering this Markov chain is that, for every fixed choice of the feature vectors, $y \mid \beta$ is now normally distributed.  Therefore, if we choose a normal prior for $\beta$, we will get that the distribution $(\beta,y)$ is jointly Gaussian, and $\beta \mid y$ is Gaussian with a tractable closed form.  These observations are enough to prove the lower bound for public-private linear regression with no distribution shift (Thm~\ref{thm:lb-linreg-no-shift-informal}), albeit with significant technical challenges involved in completing the analysis.

Finally, we consider the setting with \emph{parameter shift}.  Here, as before there are $n$ private samples and $m$ public samples.  All samples have features $\vec{x}$ chosen from $N(0,\mathbf{I}_d)$, but now we assume that for the private samples we have $y \mid \vec{x} = \langle \vec{x}, \beta_{P} \rangle + \nu$ and for the public samples we have $y \mid \vec{x} \langle \vec{x},\beta_{Q} \rangle + \nu$, for $\| \beta_{P} - \beta_{Q} \|_2 \leq \tau$.  Here we can apply a similar idea as in the case of mean estimation.  We consider the parameter shift $\beta_{Q} = \beta_{P} + v$ for a random variable $v$ that is chosen from $N(0,\nicefrac{\tau^2}{d} \cdot \mathbf{I}_d)$ so that $\| v \|_2 \leq \tau$.  Now we elide $v$ and consider the joint distribution
\begin{equation*}
(\beta_{P},y_1,\dots,y_n,y_{n+1},\dots,y_{n+m})
\end{equation*}
This distribution is jointly normal in $n+m+d$ variables and therefore the distribution $\beta_{P}$ conditioned on all the private and public data has a normal distribution with tractable mean and covariance, which we can use to complete the proof, albeit with significant technical complexities.  We overcome these technical challenges by using techniques from \emph{generalized least squares regression} where we think of the parameter shift as instead being a form of Gaussian label noise, but with a non i.i.d.\ structure, which helps us suitably modify the test statistic and complete the analysis.

\section{Preliminaries}

We briefly review the definition of differential privacy and its application in a setting with public and private data.  Let $X = (X_1,\dots,X_n) \in \gX^n$ be a dataset consisting of $n$ elements.  We say two datasets $X,X' \in \gX^n$ are \emph{neighboring} if they differ on at most one element.
\begin{definition}[Differential Privacy~\citep{dwork2006calibrating}]
    An algorithm $M : \gX^n \to \gY$ is \emph{$(\epsilon,\delta)$-differentially private} if for every pair of neighboring datasets $X,X' \in \gX^n$ and every $S \subseteq \gY$
    $$
        \Pr(M(X) \in S) \leq e^{\varepsilon} \Pr(M(X') \in S) + \delta.
    $$
\end{definition}

We consider learning algorithms that have access to both public and private data.  Specifically, we have a dataset
$$
X = (X_1,\dots,X_n,X_{n+1},\dots,X_{n+m})
$$
where we think of the first $n$ samples as \emph{private} and the last $m$ samples as \emph{public}.  In this setting we say that an algorithm $M : \gX^{n+m} \to \gY$ is \emph{$(\epsilon,\delta)$-differentially private with public and private data} if, for every possible fixed choice of the public samples $x_{n+1},\dots,x_{n+m}$, the algorithm
$$
M(X_1,\dots,X_{n},x_{n+1},\dots,x_{n+m})
$$
is $(\epsilon,\delta)$-differentially private as a function of the first $n$ elements.  Note that $M$ may depend in an arbitrary way as a function of the public samples $X_{n+1},\dots,X_{n+m}$.

In our analysis we will rely heavily on the following technical lemma about random variables that are \emph{indistinguishable} in the sense guaranteed by differential privacy.
\begin{lemma}[See e.g.\citep{narayanan2023better}]
Suppose \(0 \leq \varepsilon \leq 1\) and \(0 \leq \delta \leq \frac{1}{2}\), and that \(A, B\) are real-valued random variables such that for any set \(S\),
\[
e^{-\varepsilon} \cdot \Pr(A \in S) - \delta \leq \Pr(B \in S) \leq e^{\varepsilon} \cdot \Pr(A \in S) + \delta.
\]
Then,
\[
\left| \mathbb{E}[A - B] \right| \leq 2\varepsilon \cdot \mathbb{E}[|A|] + 2\sqrt{\delta} \cdot \mathbb{E}[A^2 + B^2].
\]
\label{prp:prob-to-exp-bound}
\end{lemma}
We defer all further definitions and technical lemmata to later in the paper where they are used.

\section{Public-Private Mean Estimation Under Distribution Shift}
\label{sec:gaussian-dist-shift}

In this section, we elaborate on details of our analysis of private mean estimation for a Gaussian distribution with unknown mean and identity covariance, given access to $n$ private samples, as well as $m$ public samples drawn from another Gaussian with a shifted mean.

First we give some notation to state our result and sketch the proof.  We have a matrix comprised of $m$ public examples $\Xpub \in \R^{m \times d}$ and $n$ private examples $\Xpriv \in \R^{n \times d}$. 
In particular, we assume that $X_{\priv, i} \sim \cal{N}(\mu_{\priv},\; \mathbf{I}_d),$ for every  $i \in [n]$,
and $X_{\pub, i} \sim \cal{N}(\mu_{\pub},\; \mathbf{I}_d),$ for every $i \in [m]$,
such that $\mu_\pub = \mu_\priv + v$ for some vector $v$ such that $\|v\|_2 \leq 2$. 
Together, $\Xpub$ and $\Xpriv$ give us  $\data$ available to the learner $M$ (where $M$ must satisfy $(\eps, \delta)$-DP with respect to the private samples). 

We split the proof of Theorem~\ref{thm:lb-gaussian-dist-shift-informal} in two parts.  The most significant part is to analyze the case where the distribution shift dominates the overall error.  That is, if we were to try using the empirical mean of the \emph{public} data $\hat\mu_{\pub} = \frac{1}{m} \sum_{i=1}^{m} X_{\pub,i}$ as an estimate of the mean of the private data, then the error would be $\| \mu_{\priv} - \hat\mu_{\pub} \|_2^2 = \tau + \nicefrac{d}{m}$, where the first term comes from the shift and the second term comes from the sampling error in the public data.  
When the sampling error dominates, the shift is largely irrelevant, so our analysis matches the case where $\tau = 0$ so there is no shift.  Our analysis for this case is captured in Theorem~\ref{thm:lb-gaussian-no-dist-shift-informal}.

\begin{theorem}[Mean estimation lower bound for $\tau=0$]
    \label{thm:lb-gaussian-no-dist-shift-informal}
    Fix any $\alpha > 0$ with $\alpha = \gO(\sqrt{d})$. Suppose $M$ is an  $(\varepsilon, \delta)$-DP learner s.t. $\E_{\data} \|M(\data) - \mu_\priv\|_2 \leq \alpha$, and $\data$ comprises of $X_\pub$ and $X_\priv$ drawn from the same distribution with mean $\mu_\priv$. When $\delta \leq \nicefrac{\varepsilon^2}{d}$, either 
    \begin{enumerate}
        \item $m={\Omega}(\nicefrac{d}{\alpha^2})$, or
        \item $n+m={\Omega}(\nicefrac{d}{\alpha \eps} + \nicefrac{d}{\alpha^2}).$
    \end{enumerate}
\end{theorem}
Note that the above theorem remains true when $\tau > 0$, as increasing $\tau$ can only make it harder to obtain an accurate estimate.  However, when $\tau \gg \sqrt{d/m}$ so that the distribution shift dominates the overall error, we want to prove a much stronger lower bound than what is in Theorem~\ref{thm:lb-gaussian-no-dist-shift-informal}, showing that no amount of public data is useful for estimating the mean of the private data.  This result is captured in Theorem~\ref{thm:lb-gaussian-dist-shift-main}.

\begin{theorem}[Mean estimation lower bound for large $\tau > 0$]
    \label{thm:lb-gaussian-dist-shift-main}
    Fix any $\alpha > 0, \tau > 0$, and $\alpha = \gO(\sqrt{d})$. Suppose $M$ is an  $(\varepsilon, \delta)$ private learner such that $\E_{\data} \|M(\data) - \mu_\priv\|_2 \leq \alpha$, and additionally
    $\delta < \nicefrac{\varepsilon^2}{d}$.
    When the shift is large, so that $\| \mu_{\mathrm{pub}} - \mu_{\mathrm{priv}} \|_2 = \tau =  \omega(\sqrt{\nicefrac{d}{m}})$, then either
    \begin{enumerate}
        \item $\tau < \alpha$, which implies $m = \Omega(d/\alpha^2)$, or
        \item $n = \Omega(\nicefrac{d}{\alpha\varepsilon} + \nicefrac{d}{\alpha^2})$
    \end{enumerate}
\end{theorem}

\begin{proof}[{\it \bfseries Proof Sketch}]
Our goal is to establish a minimax lower bound on the number of public and private samples needed for any $(\eps, \delta)$ private learner $M$ that is $\alpha$-accurate in its estimation of the private mean, \textit{i.e.}, $\|M(X) - \mu_{P}\|_2 \leq \alpha$, and for a fixed value of the shift parameter $\tau > 0$. 
Following the typical approach of minimax lower bounds, we lower bound the minimax risk with the Bayes risk under a carefully chosen prior over the unknown parameters: $\mu_P, v$, i.e., the  private mean we want to estimate and the  shift vector. 
The private mean $\mu_P$  is sampled from a spherical Gaussian $\mu_P \sim \cal{N}(0, \sigma^2 \mathbf{I}_d)$. Recall that $\mu_Q = \mu_P + v$. We define the prior over $v$ as  $v \sim \cal{N}(\mathbf{0}, \nicefrac{\tau^2}{d} \cdot \mathbf{I}_d)$. 
Thus, in expectation over the  sampling of $v$, for any value of $\mu_P$, $\E{\|\mu_P-\mu_Q\|_2^2} = \tau^2$.  The fact that there is a small probability of $\|\mu_{P} - \mu_{Q}\|_2^2 \gg \tau^2$ will not impact our proof significantly.

Next, we define our test statistic $Z_i$ for our fingerprinting method,  matching the general form in \eqref{eq:test-statistic-zi}. In particular, when $i \in [1,n]$, \textit{i.e.}, $Z_i$ tries to measure the correlation between the mechanism output $M(X)$ and a private sample $X_i$, the statistic is given by:
\[
Z_i = \innerprod{X_i - \mu_P}{M(X) - \mu_P}\quad i \in \{1\ldots n\}. 
\]
More interestingly, the statistic that measures the correlation with a public datapoint $X_j$, where $j \in \{n+1, \ldots, n+m\}$ is now re-weighted by a value that depends on the degree of shift $\tau$, \textit{i.e.},   
\[
Z_j = \frac{1}{m\cdot\nicefrac{\tau^2}{d}+1} \cdot \innerprod{X_j - \mu_P}{M(X) - \mu_P}\quad j \in \{n+1\ldots m+n\} 
\]
Essentially, when we compute the sum $\sum_i Z_i$, the weight on $Z_i$ corresponding to a public datapoint is $\Omega(1)$, when the shift is small:  $\tau = \gO(\sqrt{d/m})$, and up to constants matches the setting with no distribution shift $\tau = 0$. On the other hand, as $\tau \rightarrow \infty$, and the error from the shift dominates the sampling error over public data $\tau = \omega(\sqrt{d/m})$,  the  weight on public $Z_i$ is $o(1)$. Intuitively, we expect any accurate mechanism $M(X)$ to reduce its sensitivity over public data as the shift increases, and a similar effect is also captured by our test statistic.  We upper bound the expected sum as
\begin{equation*}
    \E\left[\sum_i Z_i\right] = \gO\paren{n\epsilon\alpha + \tfrac{1}{1 + m \cdot \nicefrac{\tau^2}{d}} \cdot \paren {\alpha \sqrt{md} + \alpha \tau \sqrt{m}}}
\end{equation*}
using differential privacy arguments we outline in Section~\ref{sec:proof-overview}. Then, we lower bound the expected value of the sum $\E[\sum_i Z_i]$ with $\Omega(d)$ when $\tau = \gO(\sqrt{d/m})$. 
For the lower bound, we use the Markov chain argument outlined in Section~\ref{sec:proof-overview}, and compute the posterior over $\mu_P \mid X$ given all the data, where the posterior mean has lower order dependence on empirical mean of the public data, as the shift grows.   
Together, with the upper bound on $\E [\sum_i{Z_i}]$, the $\Omega(d)$ lower bound gives us the final result we need on the sample complexity of $n$ and $m$ for  small values of $\tau$. For large values of $\tau$, the weight over public data points in our test statistic vanishes. In addition, we also need that $m = \Omega(d/\alpha^2)$, for the public data to estimate its own mean $\mu_Q$ to an accuracy of $\alpha$. Combining this with the condition on $\tau$, we establish that when $\alpha \gsim \tau$, only then we expect the public data to be helpful, when the shift is large ($\tau = \omega(\sqrt{d/m})$). 
\end{proof}

The the two results below together imply Theorem~\ref{thm:lb-gaussian-dist-shift-informal} in the Introduction.

\section{Public-Private Linear Regression}
\label{sec:lr-maintext}

In this section, we analyze linear regression given access to $n$ private and $m$ public samples from the \emph{same} distribution. Prior lower bounds in this setting were limited to the setting where the learner only has access to private samples~\citep{cai2021cost}.  Since the full proofs are technical, this section will only include theorem statements and a sketch of the key differneces between the proofs in this section and those in Section~\ref{sec:gaussian-dist-shift}, and the formal proofs will be deferred to the appendices.

\subsection{Lower Bounds without Distribution Shift}

First we consider the case where the public and private samples come from the same distribution.  Once again, we have $n$ private samples and $m$ public samples where $N = n+m$. In particular, we assume $N > 100d$, i.e. we have enough samples to estimate the covariance matrix. Each we assume there is some optimal linear model $\beta \in \R^d$ and we assume that each sample $(x_i, y_i)$ is drawn independently from the distribution defined by
{
\begin{align}
\label{eq:linear-regression}
y_i = x_i^\T \beta + \eta_i, 
\end{align}}where $x_i \sim \cal{N}(0, I_d)$ and $\eta_i \sim \cal{N}(0, \sigma^2)$,
and $\beta \in \R^d$ is the regression parameter of interest.
Since we will need to perform linear algebra with the data, we will use the matrix $X \in \R^{n \times d} = [ x_1, \ldots, x_N ]^\T$
and the vector $y = (y_1, \ldots, y_N)$.

\begin{theorem}
\label{thm:lr-noshift}
Let $X, y$ be generated according to the model in Equation~\ref{eq:linear-regression}. Let $M$ be an $(\eps, \delta)$-differentially private algorithm such that $\varEx{X, y}{\normtwo{M(X, y) - \beta}} < \alpha$ for some $\alpha = O(1)$. When $\delta \leq \eps^2/d$, either 
\begin{enumerate}
    \item $m = \Omega(d/\alpha^2)$, or 
    \item $n+m = \Omega(d/{\alpha\eps} + d/\alpha^2)$.
\end{enumerate}
\end{theorem}

Note that Theorem~\ref{thm:lr-noshift} corresponds directly to Theorem~\ref{thm:lb-gaussian-no-dist-shift-informal}, except for the problem of linear regression instead of mean estimation.  Specifically, it shows that when public and private samples are drawn from the same distribution, linear regression suffers the same fate as mean estimation: either the problem can be solved entirely using public samples, or we need enough samples to solve the problem privately. In the next section, we will extend this model to a shifted setting in which the private outcomes $y_i$ are generated from a model with parameter $\beta$ while the public outcomes are generated from a shifted model  with parameter $\beta + v$ for some vector $v$ such that $\|v\|_2 \leq \tau$, which corresponds to Theorem~\ref{thm:lb-gaussian-dist-shift-main}.

\begin{proof}[{\it \bfseries Proof Sketch}]
The proof follows the same high-level outline that we employ for mean estimation lower bounds.  Specifically, we will again use the Bayesian view of the fingerprinting lemma, where we consider $\beta$ sampled from some prior $\cal{N}(0, \nicefrac{1}{b}\cdot I_d)$ for some parameter $b$, and analyze the posterior distribution of $\beta$ conditioned on the public and private data.  We also modify the test statistic using the \emph{score attack} from~\citep{cai2021cost}, specifically defining
$$
Z_i = \langle \hat\beta - \beta, (y_i - x_i^\T \beta)x_i \rangle.
$$
The main technical ingredient in the proof is how we show that the test statistic is large when the estimated parameters $\hat\beta$ are close to the true parameter $\beta$, to obtain a lower bound on the error.  We condition on a fixed, typical value of the features $X$ and then compute the mean and covariance of the Gaussian posterior distribution $\beta \mid X$, which is significantly more technical than in the case of mean estimation due to the presence of inverses of random matrices.
\end{proof}

\subsection{Distribution Shift}

We now consider a distribution shift setting for linear regression.
In particular, we consider the effect of public samples
drawn from a linear model with the same covariate and noise
distribution as the private samples,
but with a shifted \emph{parameter} $\beta_\pub$.
This setting once again fits cleanly into the 
Bayesian framework we have outlined,
and we reach a similar conclusion to that in mean estimation:
when $\tau$ is small, the public data is as useful as it is
when there is no shift,
but if $\tau$ grows too large, the problem must be solved using
private data alone.

\medskip
We again observe $N=n+m$ i.i.d.\ samples, but now the public data is drawn from a linear model with a potentially different set of parameters:
\begin{align}
y_i &= x_i^{\top}\beta_{\priv}+ \eta_i, && i\in[n],\notag\\
y_i &= x_i^{\top}\beta_{\pub}+ \eta_i
      = x_i^{\top}(\beta_{\priv}+v)\;+\;\eta_i, && i\in[n+1:n+m], \label{eq:linreg-shift-model}
\end{align}
Again, we assume $x_i\sim\mathcal N(0,I_d)$ and $\eta_i\sim\mathcal N(0,\sigma^2)$ are independent.
Here we the \emph{private} parameter is $\beta_{\priv}\in\R^d$ and the \emph{public} parameter is shifted by an unknown $v \in \R^d$ such that $\normtwo{v}^2 = \tau^2$, so $\beta_{\pub}=\beta_{\priv}+v$.
The learner receives the full data set $X = \{ (x_1,y_1),\dots,(x_N,y_N)\}$ but must be $(\varepsilon,\delta)$‑DP with respect to the private rows $\{1,\dots,n\}$.

Throughout we assume $n, m > 100d$ so that the sample covariance matrices are well conditioned.  Also, as in Section~\ref{sec:lr-maintext}, the prior on the unknown regression vector is $\beta_{\priv}\sim\mathcal N\!\bigl(0,\nicefrac{1}{d} \cdot I_d\bigr)$.

The theorem below mirrors the structure of Theorem~\ref{thm:lb-gaussian-dist-shift-main} for mean estimation and of Theorem~\ref{thm:lr-noshift} for in‑distribution linear regression. As in mean estimation, for any $\tau > 0$,  the lower bound in Theorem~\ref{thm:lr-noshift} applies. Again, we are interested in the case where $\tau$ is so large that no amount of public data  will be useful and the entire burden falls on the  private samples, i.e. $\tau = \omega(\sqrt{d/m})$.

\begin{theorem}[Linear regression lower bound for large $\tau > 0$]
    \label{thm:linreg-shift}
    Fix any $\alpha > 0, \eps > 0, \tau > 0$ and assume $\alpha,\eps = O(1)$.  Suppose $M$ is an $(\eps,\delta)$-DP mechanism such that $\mathbb{E}[\normtwo{M(D) - \beta_\priv}^2] \leq \alpha^2$, and additionally $\delta < \nicefrac{\eps^2}{d}$
    When the shift is large, so that $\normtwo{\beta_\pub - \beta_\priv} = \omega(\sqrt{d/m})$, then either
    \begin{enumerate}
        \item $\tau < \alpha$, which implies $m = \Omega(d/\alpha^2)$, or
        \item $n = \Omega(\nicefrac{d}{\alpha\varepsilon} + \nicefrac{d}{\alpha^2})$.
    \end{enumerate} 

\end{theorem}

\paragraph{\textit{Proof sketch.}}
While the techniques we use for shifted linear regression are similar to those for shifted mean estimation,
the shift introduces a correlated (heteroskedastic) noise term into the linear regression model.  Specifically, if we fix the covariates $X$, then we can write the labels as Gaussian with mean $X\beta_\priv$, and label noise coming from a Gaussian with non-spherical covariance
$$\Sigma=\sigma^{2}I_N+\tfrac{\tau^{2}}{d}PP^{\top},$$
where $P=[0 , X_{\mathrm{pub}}]$ is defined from the public covariates.  That is, the distribution of the labels errors $y - X^{\top} \beta$ follows an arbitrary correlated Gaussian rather than i.i.d.\ Gaussian noise on each label.  This setting is referred to as \emph{generalized least square (GLS)} in the literature, to contrast with the more standard ordinary least squares setup, and we borrow techniques from the analysis of this GLS problem to guide our analysis.  Appendix~\ref{sec:lr-shifted} adapts the Bayesian fingerprinting framework of Section~\ref{sec:proof-overview} to the shifted linear regression setting.  To account for the non-spherical label noise, we modify the test statistic appropriately:
\begin{align}
\sum_i Z_i = \inprod{M(D) - \beta_\priv}{X^\T \Sigma^{-1}(X\beta_\priv - y)}
\end{align}
Intuitively the purpose of multiplying by $\Sigma^{-1}$ is that it restores the property that each of the terms in the inner product is distributed as a spherical Gaussian, which we relied on in the case of ordinary least squares.  The proof then follows the same general structure but with additional technical details arising from the presence of this correction.  The main feature of the analysis is that correction nicely interpolates between the cause where $\tau$ is large where the public samples are too noisy to be useful and the case where $\tau$ is small and the public and private samples are very similar.

\section{Discussion and Limitations}

In this paper, we discuss lower bounds on the sample complexity requirements for private Gaussian mean estimation and private linear regression in the presence of public data. In practice, it is common to use public data from sources that are different from the ones used to draw private data. Motivated by this, we analyze lower bounds in settings where there is a shift in the distribution of public and private samples. For this, we model the distribution shift, by allowing the public and private means, or linear regression parameters to be shifted by $\tau$ in $\ell_2$. 

Across both mean estimation and linear regression, we establish that whenever the learning objective is achievable using the combination of public and private data, one of the two sources—public or private—must be sufficient on its own whenever the distribution shift is smaller than the accuracy budget.
From a methodological standpoint, our work extends the fingerprinting lower bound technique to new regimes: (1) distribution shift in public-private mean estimation, and (2) private linear regression with and without parameter shift. By adopting a Bayesian lens on the fingerprinting method~\citep{narayanan2023better}, we simplify existing proofs and open the door to analyzing more realistic learning settings within the  learning framework of using public data to improve privacy-utility tradeoffs under differentially private learning.

\section*{Acknowledgments}
We thank Virginia Smith, Moshe Shenfeld, and Zhiwei Steven Wu for many helpful discussions early on in this work.

\bibliographystyle{plainnat}
\bibliography{pubpriv}

\appendix
\section{Private Mean Estimation Assisted with Public Data}

In this section, we analyze the hardness of private mean estimation for a Gaussian distribution with unknown mean and identity covariance. In addition to i.i.d.\ private samples, we are given i.i.d.\ public samples which can either be from the same distribution as the private samples, or from a different one. For this, we apply the Bayesian lower framework (from Section~\ref{sec:introduction}) to lower bound the expected mean of the fingerprinting statistics we will define shortly. The final lower bound on the error of any private mean estimation algorithm is then obtained by proving a tight upper bound on the expected mean of the fingerprinting statistics.  

\textbf{Setup.$\,\,$} We have a matrix comprised of $m$ public examples $\Xpub \in \R^{m \times d}$ and 
$n$ private examples $\Xpriv \in \R^{n \times d}$, where each row in them belong to the data universe $\cal{X} \subset \mathbb{R}^d$. In particular, we assume that $X_{\priv, i} \sim \cal{N}(\mu_{\priv},\; \mathbf{I}_d), \forall i \in [n]$,
and $X_{\pub, i} \sim \cal{N}(\mu_{\pub},\; \mathbf{I}_d), \forall i \in [n, n+m]$,
such that $\mu_\pub = \mu_\priv + \tau v$. When $\tau > 0$, then we are in the distribution shift setting, where the public and private data distributions observe a shifted mean, and when $\tau = 0$, then we are in the no-distribution shift setting.  
Together, $\Xpub$ and $\Xpriv$ give us the dataset $X$ available to the learner $M$.
 The learner $M$ must run private computation over the dataset $X$  and  output $M(X)$ satisfying $(\eps, \delta)$ differentially privacy with respect to the private dataset $X_\priv$. We will use  $N \eqdef n+m$ to denote the total number of examples.

Our goal is to establish a minimax lower bound on the number of public and private samples needed for any $(\eps, \delta)$ private learner $M$ that is $\alpha$ accurate in its estimation of the private mean, \textit{i.e.}, $\|M(X) - \mu_\priv\|_2 \leq \alpha$. Since minimax risk is lower bounded by the Bayes risk under any prior, we proceed by lower bounding the Bayes risk under the following prior. For a fixed and known $\tau$, the private mean $\mu_\priv$  is sampled from a spherical Gaussian $\mu_\priv \sim \cal{N}(0, \sigma^2 \mathbf{I}_d)$, and $\mu_\pub = \mu_\priv + \tau v$ where the displacement vector $v \sim \cal{N}(0,  \mathbf{I}_d)$. 

\subsection{Lower Bound for Mean-Estimation when Public and Private Distributions are Identical}

In Theorem~\ref{thm:lb-gaussian-no-dist-shift}, we present an asymptotic lower bound on the number of public and private samples needed by any $(\varepsilon, \delta)$ private learner $M$, that is $\alpha$ accurate in estimating the private mean $\mu_\priv$, in $\ell_2$ error.

\begin{theorem}[Mean estimation lower bound when public and private datapoints are i.i.d. ($\tau=0$)] 
    \label{thm:lb-gaussian-no-dist-shift} Fix any $\alpha > 0$. Suppose $M$ is an  $(\varepsilon, \delta)$ private learner that satisfies $\E_{X, \mu_\priv} \|M(X) - \mu_\priv\|_2 \leq \alpha$ for a prior distribution of $\mathcal{N}(\mathbf{0}, \sigma^2 \mathbf{I}_d)$ over $\mu_\priv$. Here, the dataset $X$ comprises of $n$ i.i.d. private samples from $\cal{N}(\mu_\priv, \mathbf{I}_d)$, and $m$ i.i.d. public samples from $\mathcal{N}(\mu_\priv, \mathbf{I}_d)$. When $\delta \leq \nicefrac{\varepsilon^2}{d}$, either $m={\Omega}(\nicefrac{d}{\alpha^2})$, or $n+m={\Omega}(\nicefrac{d}{\alpha\varepsilon} + \nicefrac{d}{\alpha^2})$.
\end{theorem}
\begin{proof} We begin by providing an overview of the proof technique via fingerprinting statistics we define shortly. 

\textbf{Overview. $\,\,$} First, we define the fingerprinting statistics, which are random variables, that depend on the randomness in the learner $M$, the sampling of the dataset $X$ and the unknown problem parameters $\mu_\priv, v$. Informally, fingerprinting statistics are random variables which cannot take very high values without violating the privacy or the accuracy of the learner. Next, we show that under appropriate priors, the sum of these fingerprinting statistics cannot be too low without violating statistical lower bounds. Together, the privacy and statistical error bounds give us the final result in Theorem~\ref{thm:lb-gaussian-no-dist-shift}. 

\textbf{Fingerprinting statistics. $\,\,$} We use $\{Z_i\}_{i=1}^N$, one for each element in the dataset (includes both private and public samples), to denote the fingerprinting statistic. 
\begin{align}
    \label{eq:zi-definition}
   Z_i = \innerprod{M(X) - \mu_\priv}{X_i - \mu_\priv}, \quad \forall i \in [N]
\end{align}
Similarly, we define the dataset $X^\prime_i$ that replaces the $i^{th}$ element in $X_\priv$ with a random sample from $\mathcal{N}(\mathbf{0}, \mu_\priv)$ if $i \in [n]$ and from $\mathcal{N}(\mathbf{0}, \mu_\pub)$ otherwise:
\begin{align}
   Z_i^\prime = \innerprod{M(X^\prime_i) - \mu_\priv}{X_i - \mu_\priv}, \quad \forall i \in [N]
\end{align}

\begin{proposition}
Assume the conditions in Theorem~\ref{thm:lb-gaussian-no-dist-shift}. Then, $\E[Z_i^\prime] = 0$ and   $\E[(Z_i^\prime)^2] = O(\alpha^2), \, i \in [n+m]$, where the expectation is taken over the sampling of $\mu_\priv$, $X$, and the randomness in the mechanism $M$. 
\label{prp:shifted-mean-zi-prime-bound}
\end{proposition}
\begin{proof}
    When conditioned on $\mu_\priv$, $X_i - \mu_\priv$ is independent of $M(X_i^\prime) - \mu_\priv$,  and $\E[X_i - \mu_\priv \mid \mu_\priv] = 0$. Thus, we have $\E[Z_i^\prime] = \E[\E[Z_i^\prime \mid \mu_\priv]] = 0$.  Next, we bound $\E[(Z_i^\prime)^2 | \mu_\priv]$ and similarly use the law of total expectation to get the upper bound on $\E[(Z_i^\prime)^2]$.
    \begin{align*}
        &\E[(Z_i^\prime)^2 | \mu_\priv] \\ 
        ={} &\E\brck{\sum_{j,k} (M(X_i^\prime) - \mu_\priv )_j (M(X_i^\prime) - \mu_\priv )_k (X_i - \mu_\priv)_j (X_i - \mu_\priv)_k \mid \mu_\priv} \\
        ={} &\sum_j \E\brck{( M(X_i^\prime) - \mu_\priv )_j^2 (X_i - \mu_\priv)_j^2 \mid \mu_\priv}  =  \E\brck{\|M(X_i^\prime) - \mu_\priv\|_2^2\mid \mu_\priv} = O(\alpha^2). 
    \end{align*}
The equality in the second step uses: (i) the conditional independence of $X_i - \mu_\priv$ and $M(X_i^\prime) - \mu_\priv$, given $\mu_\priv$, and (ii) the independence of $(X_i - \mu_\priv)_j$ and $(X_i - \mu_\priv)_j$ when $j\neq k$; and (iii) $\E[X_i - \mu_\priv \mid \mu_\priv] = 0$. The third step uses the conditional indpendence and the fact that $\E[(X_i - \mu_\priv)^2_j] = 1$. 
In the final step, we apply the assumption on the accuracy of the mechanism $M$.
\end{proof} 

\begin{proposition}
    Assume the conditions in Theorem~\ref{thm:lb-gaussian-no-dist-shift}. Then, $\E[Z_i] = O(\varepsilon \alpha), \forall i\in [n]$, where the expectation is taken over the randomness of sampling $\mu_\priv$, $X$, and the randomness in the mechanism $M$.
    \label{prp:zi-bound}
\end{proposition}
\begin{proof}
    Since $X$ and $X_i^\prime$ differ by a single element, and $M$ is $(\eps, \delta)$ differentially private, we can use the fact that the outputs $M(X)$ and $M(X_i^\prime)$ (and therefore, by the postprocessing property of DP, $Z_i$ and $Z_i'$) cannot be too different. In particular, we invoke Proposition~\ref{prp:prob-to-exp-bound}. 
\[
\left| \mathbb{E}[Z_i] - \mathbb{E}[Z'_i] \right| \leq 2\varepsilon \cdot \mathbb{E}[|Z'_i|] + 2\sqrt{\delta} \cdot \mathbb{E}[Z_i^2 + (Z'_i)^2] \leq 2(\varepsilon + \sqrt{\delta}) \cdot \sqrt{\mathbb{E}[(Z'_i)^2]} + 2\sqrt{\delta \cdot \mathbb{E}[Z_i^2]}.
\]
Next, we apply Proposition~\ref{prp:shifted-mean-zi-prime-bound}.
\[
\mathbb{E}[Z_i] \leq O((\varepsilon + \sqrt{\delta}) \alpha) + 2\sqrt{\delta \cdot \mathbb{E}[Z_i^2]}.
\]
Next, we bound $\E [Z_i^2]$ with Cauchy-Schwarz followed by the accuracy of $M$, and the concentration of $X_i$ around the private mean:
\begin{align*}
    \E [Z_i^2] \leq \;\; \sqrt{\E[\|M(X) - \mu_\priv\|_2^4]}  \sqrt{\E[\|X_i - \mu_\priv\|_2^4]} = O(\alpha^2 d)
\end{align*}
Plugging this result into the equation above, we conclude $\E[Z_i]=O(\varepsilon \alpha + \sqrt{\delta d}\alpha)$. Recall from the conditions in Theorem~\ref{thm:lb-gaussian-no-dist-shift}, $\delta \leq \nicefrac{\varepsilon^2}{d}$. Thus, $\E[Z_i] = O(\varepsilon \alpha)$.
\end{proof}

\begin{proposition}
    \label{prp:upper-bound-zi-public} Assume the conditions in Theorem~\ref{thm:lb-gaussian-no-dist-shift}.      Then, the following holds for the expected sum of the fingerprinting statistics defined for the public samples: $\E[\sum_{i=n}^{n+m}Z_i] =\gO(\alpha\sqrt{md})$.
\end{proposition}
\begin{proof}
Recall the definition of the fingerprinting statistics: $\E[Z_j] = \E\left[\inprod{M(x) - \mu_\priv}{x_j - \mu_\priv}\right]$. Then,
\begin{align*}
    \E\left[\sum_{i=n}^{n+m} Z_j\right] &= \E\left[\inprod{M(X) - \mu_\priv}{\sum_{j=n}^{n+m}(x_j - \mu_\priv)}\right]\\
    &\leq \sqrt{\E[\norm{M(X)-\mu_\priv}{2}^2] \cdot \sum_{j=n}^{n+m}\E[ \norm{x_j - \mu_\priv}{2}^2]} \\
    & \leq \alpha  \sqrt{\sum_{j=n}^{n+m}\E\brck{\E[\norm{x_j - \mu_\priv}{2}^2 \mid \mu_\priv]}} = \gO(\alpha \sqrt{md}).
\end{align*}
The first inequality uses Cauchy-Schwarz, and the second uses the bound on the accuracy of the mechanism $M$. Finally, the last equality uses the tower law of expectations, followed by the variance of the public data around the public mean $\mu_\pub$, which matches $\mu_\priv$ in the no distribution shift setting.  
\end{proof}
\begin{corollary}
    \label{corr:sum-zi-upper-bound}
    Under the conditions of Theorem~\ref{thm:lb-gaussian-no-dist-shift}, for the expectation taken over the randomness of $\mu_\priv, X_\priv$ and the mechanism $M$, $\sum_{i=1}^{n+m} \E[Z_i] = O(n\varepsilon\alpha + \alpha\sqrt{md})$.
\end{corollary}

Next, we lower bound $\sum_{i=1}^{n+m} Z_i = (n+m) \innerprod{M(X) - \mu_\priv}{\bar{\mu} - \mu_\priv}$, where $\bar{\mu} = \nicefrac{1}{n+m} \sum_{i=1}^{n+m} Z_i$, \textit{i.e.}, it is the empirical mean of the public and private samples combined. It is sufficient to lower bound $\innerprod{M(X) - \mu_\priv}{\bar{\mu} - \mu_\priv}$.

\begin{lemma}
 For $n+m>0$, we can lower bound $\innerprod{M(X) - \mu_\priv}{\bar{\mu} - \mu_\priv}$, in the following way:
\begin{align}
\mathbb{E}\big[\innerprod{M(X) - \mu_\priv}{\bar{\mu} - \mu_\priv} \big] 
&\geq \mathbb{E}\big[\|\mu_\priv - \bar{\mu}\|_2^2\big] - \label{eq:main-breakdown}\sqrt{\mathbb{E}\big[\|M(X) - \bar{\mu}\|_2^2\big] \mathbb{E}_X\big[\| \mathbb{E}[\mu_\priv \mid X] - \bar{\mu}\|_2^2\big]}.
\end{align}
\label{lemma:breakdown}
\end{lemma}
\begin{proof}
First we add and subtract the empirical mean from the first term:
\begin{equation}
    \label{eq:breakdown-one}
    \innerprod{M(X) - \mu_\priv}{\bar{\mu} - \mu_\priv} = \innerprod{M(X) - \bar{\mu}}{\bar{\mu} - \mu_\priv} + \|\bar{\mu} - \mu_\priv\|_2^2.
\end{equation} 
Next, we bound the magnitude of $\E\brck{\innerprod{M(X) - \bar{\mu}}{\bar{\mu} - \mu_\priv}}$, and for this we note that conditioned on the data $X$, the randomness in the mechanism $M$ is independent of $\mu_\priv$. This conditional independence trick allows us to break the expectation down as follows:
\begin{align*}
    \abs{\E\brck{\innerprod{M(X) - \bar{\mu}}{\bar{\mu} - \mu_\priv}}} &= \E \brck{\E\brck{\innerprod{M(X) - \bar{\mu}}{\bar{\mu} - \mu_\priv}} \mid X} \\
    &=\E\brck{\innerprod{\E\brck{M(X) - \bar{\mu} \mid X)}}{\E\brck{\bar{\mu} - \mu_\priv\mid X}}} \\
    &\leq \sqrt{\E_X\brck{\|\E\brck{M(X) - \bar{\mu} \mid X)}\|_2^2}} \cdot \sqrt{\E_X\brck{\|\E\brck{\bar{\mu} - \mu_\priv\mid X}\|_2^2}},
\end{align*}
where the last inequality uses Cauchy-Schwarz. Applying Jensen inequality on the first term we get the desired lower bound on $\E\brck{\innerprod{M(X) - \bar{\mu}}{\bar{\mu} - \mu_\priv}}$:
\begin{align*}
    \E\brck{\innerprod{M(X) - \bar{\mu}}{\bar{\mu} - \mu_\priv}} \geq -  \sqrt{\mathbb{E}\big[\|M(X) - \bar{\mu}\|_2^2\big] \cdot \mathbb{E}_X\big[\| \mathbb{E}[\mu_\priv \mid X] - \bar{\mu}\|_2^2\big]}.
\end{align*}
Plugging the above inequality into $\eqref{eq:breakdown}$ completes the proof. 
\end{proof}

\begin{lemma}
    \label{lemma:posterior-concentration}
    Consider the setup in Theorem~\ref{thm:lb-gaussian-no-dist-shift}.  , Then, $
        \E\brck{\|\E[\mu_\priv \mid X] - \bar{\mu}\|_2^2}  = \gO\paren{\frac{d}{(n+m)^3}}.$
\end{lemma}
\begin{proof}
    Follows from $gN(\mathbf{0}, \sigma^2 \mathbf{I}_d)$ being a conjugate prior over the private mean $\mu_\priv$, and the fact that $\bar{\mu} \sim \gN(\mu_\priv, \nicefrac{1}{n+m}\cdot \mathbf{I}_d)$. The mean for the posterior distribution over $\mu_\priv \mid X$ is given by $\bar{\mu}\cdot \nicefrac{(n+m)\sigma^2}{(n+m)\sigma^2 + 1}$. Thus,
    \begin{align*}
        \E\brck{\|\E[\mu_\priv \mid X] - \bar{\mu}\|_2^2}  = \E\brck{\|\bar{\mu}\|_2^2} \cdot \frac{1}{((n+m)\sigma^2 + 1)^2} = \gO\paren{\frac{d}{(n+m)^3}}.
    \end{align*}
The last equality follows from the expected norm of $\bar{\mu}$ computed over  $X, \mu_\priv$. Note that we can bound $\E\|\bar{\mu}\|_2 = \E_{\mu_\priv} [\E_{X} [ \|\bar{\mu}\|_2 \mid  \mu_\priv] ] = \gO(\sqrt{\nicefrac{d}{(n+m)}})$ since $\E_{X} [ \|\bar{\mu}\|_2 \mid  \mu_\priv] = \gO(\sqrt{\nicefrac{d}{(n+m)}})$~\cite{wainwright2019high}. 
\end{proof}

\begin{lemma}
    \label{lemma:statistical-lb}
    Under the conditions of Theorem~\ref{thm:lb-gaussian-no-dist-shift}, where $\E\brck{\|M(X) - \mu_\priv\|_2} \leq \alpha$, the number of total data points $n+m = \Omega(\nicefrac{d}{\alpha^2})$.
\end{lemma}
\begin{proof}
    Recall that for any random vector \( v \) with mean \( \mu \), $\mathbb{E}\left[\|v\|^2\right] = \mathbb{E}\left[\|v - \mu\|^2\right] + \|\mu\|^2 \geq \mathbb{E}\left[\|v - \mu\|^2\right]$. Conditioned on the dataset $X$, $M(X) - \mu_\priv$ is a random sample from distribution with mean $M(X) - \E[\mu_\priv \mid X]$. Applying the above inequality on this vector gives us:
\begin{align*}
    \E\|\mu_\priv - M(X)\|_2^2 \geq \E_X \brck{\|\E \brck{ \mu_\priv \mid  X} - M(X)\|_2^2}. 
\end{align*}
Also, note that:
\begin{align*}
    \E\brck{\|\mu_\priv - \bar{\mu}\|_2^2} \leq 2\paren{\E\brck{\|\E \brck{ \mu_\priv \mid  X} - \mu_\priv\|_2^2} + \E\brck{ \| \E \brck{ \mu_\priv \mid  X}  - \bar{\mu}\|_2^2}},
\end{align*}
which implies that $\E\|\mu_\priv - M(X)\|_2^2 \geq  \E\brck{\|\mu_\priv - \bar{\mu}\|_2^2} - 2 \E\brck{ \| \E \brck{ \mu_\priv \mid  X}  - \bar{\mu}\|_2^2}$. From Lemma~\ref{lemma:posterior-concentration}, we know that $\E\brck{ \| \E \brck{ \mu_\priv \mid  X}  - \bar{\mu}\|_2^2} = \gO(\nicefrac{d}{(n+m)^3})$. Since $\E\brck{\|\mu_\priv - \bar{\mu}\|_2^2}  = \Omega(\nicefrac{d}{(n+m)})$~\cite{vershynin2018high}, we conclude that:  $\E\|\mu_\priv - M(X)\|_2^2 = \Omega(\nicefrac{d}{(n+m)})$. Thus, for $\alpha^2 \geq \E\|\mu_\priv - M(X)\|_2^2 $, we need $n+m = \Omega(\nicefrac{d}{\alpha^2})$.
\end{proof}

\begin{corollary}
    \label{corr:sum-zi-lower-bound}
    Under the conditions of Theorem~\ref{thm:lb-gaussian-no-dist-shift}, when $\alpha = \gO(\sqrt{d})$, for the expectation taken over the randomness of $\mu_\priv, X_\priv$ and the mechanism $M$, $\sum_{i=1}^{n+m} \E[Z_i] = \Omega(d)$.
\end{corollary}
\begin{proof}
    We can write $\sum_{i=1}^{n+m} Z_i = (n+m) \cdot \mathbb{E}\big[\innerprod{M(X) - \mu_\priv}{\bar{\mu} - \mu_\priv} \big]$, and from Lemma~\ref{lemma:breakdown} we know:
    \begin{align}
        \mathbb{E}\big[\innerprod{M(X) - \mu_\priv}{\bar{\mu} - \mu_\priv} \big] 
&\geq \mathbb{E}\big[\|\mu_\priv - \bar{\mu}\|_2^2\big] 
\nonumber - \sqrt{\mathbb{E}\big[\|M(X) - \bar{\mu}\|_2^2\big] \cdot \mathbb{E}_X\big[\| \mathbb{E}[\mu_\priv \mid X] - \bar{\mu}\|_2^2\big]}.
    \end{align}
    Next, we apply the result $\mathbb{E}\big[\|\mu_\priv - \bar{\mu}\|_2^2\big] = \Theta(\nicefrac{d}{m+n})$, purely from known upper and lower bounds for the concentration of the empirical mean, when samples are drawn i.i.d. from $\gN(\mathbf{0}, \mathbf{I}_d)$~\cite{wainwright2019high}.
    Similarly, we can bound $\mathbb{E}\big[\|M(X) - \bar{\mu}\|_2^2\big]  = \mathbb{E}\big[\|M(X) - \mu_\priv\|_2^2\big] + \mathbb{E}\big[\|\mu_\priv - \bar{\mu}\|_2^2\big] = \gO(\alpha^2 + \nicefrac{d}{n+m})$. Finally, we apply the Lemma~\ref{lemma:posterior-concentration} on the concentration of the posterior mean, around $\mu_\priv$ to get:
    \begin{align}
        \mathbb{E}\big[\innerprod{M(X) - \mu_\priv}{\bar{\mu} - \mu_\priv} \big] =  
&\Theta\paren{\frac{d}{n+m}} - \gO\paren{\sqrt{\paren{\alpha^2 + \frac{d}{n+m}} \cdot \frac{d}{(n+m)^3} }}.
    \end{align}
Now plugging in the condition that $\alpha = \gO(\sqrt{d})$, we get $\sum_{i=1}^{n+m}\E[Z_i] = \Omega(d)$
\end{proof}

We are now ready to prove the main result in Theorem~\ref{thm:lb-gaussian-dist-shift}.  From Corollary~\ref{corr:sum-zi-upper-bound} and Corollary~\ref{corr:sum-zi-lower-bound}, there exists constants $c_1, c_2 > 0, n_0>0$, such that for all $n \geq n_0, \delta \leq \nicefrac{\varepsilon^2}{d}$, the following is true:
\begin{align*}
 c_1 d \; \leq \;  \E \brck{\sum_{i=1}^{m+n} Z_i} \; \leq \; c_2\paren{n\varepsilon\alpha + \alpha \sqrt{md}}.
\end{align*}
The above implies that either of the following conditions is true. First,  $n\varepsilon\alpha = \Omega(d)$ or $n = \Omega(\nicefrac{d}{\varepsilon\alpha})$. Second, $\alpha\sqrt{md} = \Omega(d)$ or $m = \Omega(\nicefrac{d}{\alpha^2})$. Combining this result with Lemma~\ref{lemma:statistical-lb}, we conclude that either $m = \Omega(\nicefrac{d}{\alpha^2})$ or $n + m  = \Omega(\nicefrac{d}{\alpha\varepsilon} + \nicefrac{d}{\alpha^2})$. We can interpret this result as the following: either we need enough public samples to estimate the mean $\mu_\priv$ to accuracy $\alpha$ without the help of any private data, or we must have enough public and private data combined, to be able to treat the public data as private, and estimate the accuracy of $\mu_\priv$ to accuracy $\alpha$.  
\end{proof}

\subsection{Lower Bound for Mean Estimation when Public Distribution is Shifted from Private}
\label{subsec:distribution-shift-lower-bound}

First, we re-state a formal version of our Gaussian mean estimation result under distribution shift (Theorem~\ref{thm:lb-gaussian-dist-shift-main}). Then we provide a proof overview, that defines the test-statistic,  followed by a presentation of the upper and lower bounds on the expected sum of the test-statistics.

\begin{theorem}[Lower bound on public-assisted private Gaussian mean estimation when public \& private data distributions differ]
    \label{thm:lb-gaussian-dist-shift}
    Fix any $d>0$, $10\sqrt{d} > \alpha > 0$ and $\tau > 0$. Suppose $M$ is an  $(\varepsilon,\delta)$ private learner, with $\delta \leq \nicefrac{\varepsilon^2}{d}$ that takes as input dataset $X$ and satisfies an expected accuracy of $\E_{X} \|M(X) - \mu_\priv\|_2 \leq \alpha$. 
    Here, the dataset $X$ available to $M$ comprises of $n$ private samples $X_1, \ldots, X_n \sim \mathcal{N}(\mu_\priv, \mathbf{I}_d)$ and $m$ public samples $X_{n+1}, \ldots, X_{n+m} \sim \mathcal{N}(\mu_\pub, \mathbf{I}_d)$, where the public and private distributions differ, i.e., where $\mu_\pub = \mu_\priv +  v$ and $\normtwo{v} = \tau$. Then, the following is true: 
    \begin{itemize}
        \item when $\tau = \gO(\nicefrac{d}{\sqrt{m}})$, then either $m = \Omega(\nicefrac{d}{\alpha^2})$ or $n+m = \Omega(\nicefrac{d}{\alpha\varepsilon} + \nicefrac{d}{\alpha^2})$,
        \item and when $\tau = \omega(\sqrt{\nicefrac{d}{m}})$, then either $\alpha \gsim  \tau$ or  $n = \Omega(\nicefrac{d}{\alpha\varepsilon} + \nicefrac{d}{\alpha^2})$. 
    \end{itemize} 
\end{theorem}
\textbf{Proof overview.} As with any minimax lower bound, we step into the lower bound calculations by constructing a prior over $\mu_\priv, v$, and then computing the lower bound on sample complexity of $n, m$ in expectation  over this prior. The prior we choose over $\mu_\priv$ is similar to the no distribution shift with $\mu_\priv \sim \gN(\mathbf{0}, \mathbf{I}_d)$, and $v \sim \gN(\mathbf{0}, \nicefrac{\tau^2}{d} \cdot \mathbf{I}_d)$. Next, we construct our fingerprinting statistics that measure the correlation between mechanism output $M(X)$ and any private data point $X_i$. This test-statistic cannot be too high, else we violate differential privacy guarantee, and cannot be too low, else accuracy cannot be guaranteed, since if it is too low, it means that that the mechanism affords very low sensitivity on any given data point,. 

\textit{Remark on how distribution shift affects thee design of test-statistic.} As the distribution shift gets more severe, i.e., as $\tau$ increases, we expect public data to be less useful for any accurate mechanism $M$. Thus, we expect the correlation between $M(X)$, and any public data point $X_j$ to get lower, i.e., more generally for any test-statistic that measures the presence of a private data point $X_i$ in the  dataset $X$ vs. not, the sensitivity of the test- statistic should reduce as $\tau$ increases. Motivated by this, we define the following test-statistic $Z_i$ for the Gaussian mean estimation with distribution shift setting:
\begin{align}
    Z_i &=  \innerprod{M(X) - \mu_\priv}{X_i - \mu_\priv}, \quad i \in \{1,\ldots, n\}~~ (\text{private } X_i) \nonumber\\
    Z_i &= \frac{1}{\nicefrac{m\tau^2}{d}+1} \cdot  \innerprod{M(X) - \mu_\priv}{X_i - \mu_\priv},  \quad  i \in \{n+1,\ldots, n+m\} ~~ (\text{public } X_i) \label{eq:test-statistic-dist-shift}
\end{align}

Similar to our proof of the lower bound in the identical distribution setting, we upper and lower bound the expected sum of the fingerprinting statistics above. 
We begin with the upper bound, on $\E[Z_i]$ in \eqref{eq:test-statistic-dist-shift}, and we bound the sum of this statistic separately for the public and private samples in Proposition~\ref{prp:upper-bound-zi-dist-shift}.  
\begin{proposition}
    \label{prp:upper-bound-zi-dist-shift} Assume the conditions in Theorem~\ref{thm:lb-gaussian-dist-shift}.  The expected sum of the fingerprinting statistics corresponding to private samples satisfies:  $\E[\sum_{i=1}^{n}Z_i] = \gO(n \varepsilon\alpha)$, and the one corresponding to public samples satisfies: $\sum_{i=n+1}^{n+m}  \E[Z_i] =  \gO(\alpha\sqrt{md} + \alpha\tau\sqrt{md})$.
\end{proposition}
\begin{proof}
The proof for the upper bound over $\E[\sum_{i=1}^{n}Z_i]$ is similar to  Proposition~\ref{prp:zi-bound},  which uses the DP constraint to show $\E[\sum_{i=1}^{n}Z_i] = \gO(n \varepsilon\alpha)$. For the second term which sums the fingerprinting statistics corresponding to public samples, $\E[\sum_{i=n+1}^{n+m}Z_i]$ we need to additionally account for the randomness in the direction of shift $v$, when computing the expectation over the public mean $\bar{\mu}_\pub$. 
Recall the definition of the fingerprinting statistics: $Z_j = \inprod{M(X) - \mu_\priv}{x_j - \mu_\priv}$. Then,
\begin{align*}
    \E\left[\sum_{j=n+1}^{n+m} Z_j\right] &= \E\left[\inprod{M(X) - \mu_\priv}{\sum_{j=n+1}^{n+m}(x_j - \mu_\priv)}\right]\\
    &\leq \sqrt{\E[\norm{M(X)-\mu_\priv}{2}^2]}\sqrt{\sum_{j=n+1}^{n+m}\E[ \norm{x_j - \mu_\priv}{2}^2]} 
\end{align*}
\begin{align*}
     \E\left[\sum_{j=n+1}^{n+m} Z_j\right]  & \leq \alpha  \sqrt{\sum_{j=n+1}^{n+m}\E\brck{\E[\norm{x_j - \mu_\priv}{2}^2 \mid \mu_\priv]}}\\
    &= \alpha  \sqrt{\sum_{j=n+1}^{n+m}\E\brck{\E[\|x_j - \mu_\priv - v \mid \mu_\priv\|^2_2]} + \E[\|v\|_2^2]} = \gO(\alpha \sqrt{md} + \alpha\tau\sqrt{md}).
\end{align*}
The first inequality uses Cauchy-Schwarz, and the second uses the bound on the accuracy of the mechanism $M$. Finally, the last equality uses the tower law of expectations, followed by the indpendence between displacement $v$ and private mean  $\mu_\priv$.  Then, we simply apply the definitions for the priors over public and private means to get the final result.  
\end{proof}

\begin{corollary}
    \label{corr:sum-zi-upper-bound-dist-shift}
    Under the conditions of Theorem~\ref{thm:lb-gaussian-dist-shift}, over the randomness of $\mu_\priv, X, v$ and mechanism $M$, $ \sum_{i=1}^{n} \E[Z_i] + \frac{1}{\nicefrac{m}{d}\cdot\tau^2 + 1} \cdot \sum_{i=n+1}^{n+m} \E[Z_i] = O\paren{n\varepsilon\alpha + \frac{1}{\nicefrac{m}{d}\cdot\tau^2 + 1} \cdot \paren{\alpha\sqrt{md} + \alpha\tau\sqrt{md}}}$.
\end{corollary}
\begin{proof}
    Follows immediately from Proposition~\ref{prp:upper-bound-zi-dist-shift} that upper bounds the second term in the summation and Proposition~\ref{prp:zi-bound} which upper bounds the first term.
\end{proof}

Next, we lower bound the expected sum over the fingerprinting statistics, for the expression:
\[
\sum_{i=1}^{n} \E[Z_i] + \frac{1}{\nicefrac{m}{d}\cdot\tau^2 + 1} \cdot \sum_{i=n+1}^{n+m} \E[Z_i].
\]
Our approach for the lower bound is similar to the case where there is no shift, but we explicitly account for the distribution shift term $v$ in our posterior calculations. Recall that our prior for $v$ is $v \sim \gN(0, \tfrac{\tau^2}{d}\cdot I_d)$. 
We will see that when $\tau = \gO(\sqrt{d/m})$, the lower bound on the above expression scales as the one where $\tau = 0$, i.e., when there is no shift.

\begin{lemma}
    \label{lemma:posterior-computation-dist-shift}
    Consider the setup in Theorem~\ref{thm:lb-gaussian-dist-shift}. Then, $\E[\mu_\priv \mid X] = w_\priv \cdot \bar{\mu}_\priv + w_\pub \cdot \bar{\mu}_\pub$, where $w_\priv = \frac{\sigma^2 (\tau^2/d + \nicefrac{1}{m})}{\sigma^2 (\tau^2/d + \nicefrac{1}{m} + \nicefrac{1}{n}) + \nicefrac{\tau^2}{dn} + \nicefrac{1}{mn}}$ and $w_\pub = \frac{\nicefrac{\sigma^2}{n}}{\sigma^2 (\nicefrac{\tau^2}{d} + \nicefrac{1}{m} + \nicefrac{1}{n}) + \nicefrac{\tau^2}{nd} + \nicefrac{1}{mn}}$. 
\end{lemma}
\begin{proof}
    Let $\bar{\mu}_\pub$ be the empirical mean of the public data and $\bar{\mu}_\priv$ be the empirical mean of the private data points. 
    We can then derive the following joint distribution over $\mu_\priv$, $\bar{\mu}_\pub, \bar{\mu}_\priv$:
\begin{align}
& (\mu_\priv, \bar{\mu}_\pub, \bar{\mu}_\priv) \sim \mathcal{N}\left(0, \Sigma\right), \; \Sigma \in \mathbb{R}^{3d \times 3d} \;\;\;\;\; \text{and,} \label{eq:joint-distribution} \\ \Sigma &= \begin{pmatrix}
\sigma^2 \mathbf{I}_d & \sigma^2 \mathbf{I}_d & \sigma^2 \mathbf{I}_d \\
\sigma^2 \mathbf{I}_d & (\sigma^2 + \nicefrac{\tau^2}{d} + \nicefrac{1}{m}) \cdot \mathbf{I}_d & \sigma^2 \mathbf{I}_d\\
\sigma^2 \mathbf{I}_d & \sigma^2 \mathbf{I}_d & (\sigma^2 + \nicefrac{1}{n}) \cdot\mathbf{I}_d, \nonumber
\end{pmatrix}.
\end{align}

and $\E[\mu_\priv \mid \bar{\mu}_\pub, \bar{\mu}_\priv]$ is given by $ \Sigma_{12}\Sigma_{22}^{-1}\paren{[\bar{\mu}_\pub, \bar{\mu}_\priv]^\top - [\E[\mu_\priv], \E[\mu_\priv]]^\top}$, where $\Sigma_{12}$, and $\Sigma_{22}$ are given by the top $d \times 2d$, and bottom $2d \times 2d$ submatrices of $\Sigma$. Based, on this, we get:
\begin{align*}
\Sigma_{12}\Sigma_{22}^{-1} = \left( 
    \frac{\sigma^2 / n}{(\sigma^2 + \tau^2/d + 1/m)(\sigma^2 + 1/n) - \sigma^4} I_d, \frac{\sigma^2(\tau^2/d + 1/m)}{(\sigma^2 + \tau^2/d + 1/m)(\sigma^2 + 1/n) - \sigma^4} I_d\right)
\end{align*}
Then, the mean of the posterior distribution over $\mu_\priv$ is given by:
\begin{align*}
\E[\mu_\priv \mid \bar{\mu}_\pub, \bar{\mu}_\priv] &= \left( 
    \frac{\sigma^2 / n}{(\sigma^2 + \tau^2/d + 1/m)(\sigma^2 + 1/n) - \sigma^4} I_d\right)\cdot(\bar{\mu}_{\pub} - \E[\mu_{\pub}]) \\    & \quad\quad +\left(\frac{\sigma^2(\tau^2/d + 1/m)}{(\sigma^2 + \tau^2/d + 1/m)(\sigma^2 + 1/n) - \sigma^4} I_d\right) \cdot(\bar{\mu}_{\priv} - \E[\mu_{\priv}]) \\
    &= \frac{\nicefrac{\sigma^2}{n} \cdot \bar{\mu}_\pub + \sigma^2 \bar{\mu}_\priv \cdot (\tau^2/d + \nicefrac{1}{m})} {\sigma^2 (\tau^2/d + \nicefrac{1}{m} + \nicefrac{1}{n}) + \nicefrac{\tau^2}{nd} + \nicefrac{1}{mn}}
\end{align*}
From the expression above it it easy to derive $w_\pub$ and $w_\priv$ in the statement of Lemma~\ref{lemma:posterior-computation-dist-shift}.
\end{proof}

\begin{lemma}
    \label{lemma:gaussian-posterior-mean-diff}
    Under the conditions in Theorem~\ref{thm:lb-gaussian-dist-shift},
    \begin{align*}
    \mathbb{E}_X \|\E\brck{{\mu_\priv\mid X}}-\bar{X}\|_2^2 =  \gO\paren{\frac{d}{\paren{\nicefrac{m}{\kappa}+n}^3} },~~~\text{where}~~\kappa\eqdef \frac{m\tau^2}{d} + 1,
    \end{align*}
    and the inner expectation is taken over $\mu_\priv, v \mid X$, and $\bar{X}$ is defined as:
    \begin{align}
    \bar{X} \eqdef \bar{X}_\pub \cdot \frac{\nicefrac{m}{(m\tau^2/d +1)}}{n+\nicefrac{m}{(m\tau^2/d +1)}} + \bar{X}_\priv \cdot \frac{n}{n+\nicefrac{m}{(m\tau^2/d +1)}}, 
\end{align}
with $\bar{X}_\pub$ and $\bar{X}_\priv$ being the empirical mean over public and private data points.  
\end{lemma}

\begin{proof}

Directly applying Lemma~\ref{lemma:posterior-computation-dist-shift} we get:
\begin{align}
    &\E\brck{\mu_\priv \mid X} - \bar{X}    \nonumber \\
     ={}  &\frac{\nicefrac{\sigma^2}{n} \cdot \bar{X}_\pub + \sigma^2 \bar{X}_\priv \cdot (\tau^2/d + \nicefrac{1}{m})} {\sigma^2 (\tau^2/d + \nicefrac{1}{m} + \nicefrac{1}{n}) + \nicefrac{\tau^2}{nd} + \nicefrac{1}{mn}} - \frac{n}{\nicefrac{m}{(m\tau^2/d+1)}+n}\cdot\bar{X}_\priv  - \frac{\nicefrac{m}{(m\tau^2/d+1)}}{\nicefrac{m}{(m\tau^2/d+1)}+n}\cdot
    \bar{X}_\pub\label{eq:breakdown-5} \\
    ={} &\bar{X}_\pub \paren{ \frac{\nicefrac{\sigma^2}{n}}{\sigma^2 (\tau^2/d + \nicefrac{1}{m} + \nicefrac{1}{n}) + \nicefrac{\tau^2}{nd} + \nicefrac{1}{mn}} - \frac{\nicefrac{m}{(m\tau^2/d+1)}}{\nicefrac{m}{(m\tau^2/d+1)}+n}} \nonumber \\
    &+ \bar{X}_\priv \paren{ \frac{\sigma^2(\tau^2/d + 1/m)}{\sigma^2 (\tau^2/d + \nicefrac{1}{m} + \nicefrac{1}{n}) + \nicefrac{\tau^2}{nd} + \nicefrac{1}{mn}}- \frac{n}{\nicefrac{m}{(m\tau^2/d+1)}+n}}. \label{eq:breakdown-6}
\end{align}

To bound the expected squared norm of the above term, we first need to bound terms $\alpha, \beta$ defined  as follows:
\[
\alpha \eqdef \frac{\nicefrac{\sigma^2}{n}}{\sigma^2 (\tau^2/d + \nicefrac{1}{m} + \nicefrac{1}{n}) + \nicefrac{\tau^2}{nd} + \nicefrac{1}{mn}} - \frac{\nicefrac{m}{(m\tau^2/d+1)}}{\nicefrac{m}{(m\tau^2/d+1)}+n}
\]
\[
\beta \eqdef   \frac{\sigma^2(\tau^2/d + 1/m)}{\sigma^2 (\tau^2/d + \nicefrac{1}{m} + \nicefrac{1}{n}) + \nicefrac{\tau^2}{nd} + \nicefrac{1}{mn}} - \frac{n}{\nicefrac{m}{(m\tau^2/d+1)}+n}
\]

Let $\sigma^2 = 1$ and define $\kappa := 1 + \frac{m \tau^2}{d}$. Then we can simplify in the following way:

\begin{align}
    \alpha = \frac{-m\kappa}{(n \kappa + \kappa +m)(n\kappa+m)}, 
\quad
    \beta = \frac{-n\kappa^2}{(n \kappa + \kappa +m)(n\kappa+m)} 
\end{align}

Thus, $\alpha \bar{X}_\pub + \beta \bar{X}_\priv$ can be written as:
\begin{align}
    \alpha \bar{X}_\pub + \beta \bar{X}_\priv = -\frac{\paren{\kappa^2 \sum_{i=1}^n X_i + \kappa \sum_{j=n+m}^{n+1} X_j}}{(m+n\kappa+\kappa)(m+n\kappa)} \label{eq:1001}
\end{align}

We can bound the $\norm{\cdot}{2}^2$ for the expression, by noting that $\norm{\sum_{i=1}^n X_i }{2}^2 = \gO(nd)$ and $\norm{\sum_{i=m+1}^{n+m} X_i }{2}^2 = \gO(md)$. Applying this in the above expression, and dividing both numerator and denominator by $\kappa^2$ we get the final result:
\begin{align}
    \norm{\E\brck{\mu_\priv \mid X} - \bar{X}    \nonumber}{2}^2  &= \gO\paren{\frac{d(n+\nicefrac{m}{\kappa})}
        {
    \paren{\nicefrac{m}{\kappa}+n+1}^2\paren{\nicefrac{m}{\kappa}+n}^2}}  \\
    &= \gO\paren{\frac{d}{\paren{n+\nicefrac{m}{\kappa}}^3}}
\end{align}
This completes the proof of Lemma~\ref{lemma:gaussian-posterior-mean-diff}.
\end{proof}

\begin{lemma}
    \label{lemma:gaussian-lb-priv}
    Under the conditions in Theorem~\ref{thm:lb-gaussian-dist-shift}, the expected sum of the fingerprinting statistics is lower bounded as follows:
    \begin{align*}
    \mathbb{E} \brck{ \sum_i Z_i } =  \Omega\paren{(n+\nicefrac{m}{\kappa}) \cdot \paren{\frac{d}{n+\nicefrac{m}{\kappa}} - \sqrt{\paren{\alpha^2 + \frac{d}{n + \nicefrac{m}{\kappa}}} \cdot \paren{\frac{d}{(n+\nicefrac{m}{\kappa})^3} }}}},
    \end{align*}
    where $\kappa = (m\tau^2/d+ 1)$ and the expectation is taken  over the randomness of $\mu_\priv$, $X$, and the mechanism $M$. 
\end{lemma}
\begin{proof}
    Similar to our previous proofs on the lower bound for the expected sum of test statistics,  we manipulate the following sum into two terms. 
\begin{align}
    &\E_{\mu_\priv, X, v}\left[\sum_{i=1}^{n} Z_i + \frac{1}{\nicefrac{m}{d}\cdot\tau^2 + 1} \cdot \sum_{i=n+1}^{n+m} Z_i \right] \nonumber \\
    &= (n+\nicefrac{m}{(m\tau^2/d+1)}) \cdot \left(\E\left[\|\bar{X}-\mu_\priv\|_2^2\right] + \E \left\langle M(X)-\bar{X}, \bar{X}-\mu_\priv\right\rangle \right), \label{eq:breakdown}
\end{align}

Given, $X$, the random variable $M(X)$ is independent of $\mu_\priv$. 
\begin{align}
    \label{eq:breakdown-2}
    \E_{\mu_\priv, X, v}  &  \brck{ \left\langle M(X)-\bar{X}, \bar{X}-\mu_\priv\right\rangle } \;\; =\;\; \E_X \brck{\innerprod{\E_{\mu_\priv, v \mid X}[M(X)-\bar{X}]}{\E_{\mu_\priv, v \mid X}[\bar{X} -  \mu_\priv]}} \nonumber\\
 & \;\; \leq \;\;  \sqrt{\E_X\|\E_{\mu_\priv, v \mid X}[M(X)-\bar{X}]\|_2^2} \cdot \sqrt{\E_X\|\E_{\mu_\priv, v \mid X}[\bar{X} -  \mu_\priv]\|_2^2},
\end{align}
where the inequality uses Cauchy-Schwarz. We can further use the convexity of $\|\cdot\|_2^2$ in the first term above to arrive at the following upper bound.
\begin{align}
    \label{eq:breakdown-4}
    & \E_{\mu_\priv, X, v} \brck{ \left\langle M(X)-\bar{X}, \bar{X}-\mu_\priv\right\rangle}  \nonumber\\
 & \;\; \leq \;\;  \sqrt{\E_{\mu_\priv, v , X} \|[M(X)-\bar{X}]\|_2^2}  \cdot \sqrt{\E_X\|\E_{\mu_\priv, v \mid X}[\mu_\priv] -  \bar{X}\|_2^2},
\end{align}
Next, we proceed by bounding $\E_{\mu_\priv, v , X} \|[M(X)-\bar{X}]\|_2^2$.
\begin{align}
    \E_{\mu_\priv, v , X} \|[M(X)-\bar{X}]\|_2^2 &=   \E_{\mu_\priv, v , X} \|\mu_\priv-\bar{X}]\|_2^2 + \E_{\mu_\priv, v , X} \|[M(X)-\mu_\priv]\|_2^2 \nonumber \\
   & \leq  \E_{\mu_\priv, v , X} \|\mu_\priv-\bar{X}\|_2^2 + \alpha^2 \label{eq:breakdown-8-2}
\end{align}
Recall the definition of $\bar{X}$: 
\begin{align}
    \bar{X} \eqdef \bar{X}_\pub \cdot \frac{\nicefrac{m}{(m\tau^2/d +1)}}{n+\nicefrac{m}{(m\tau^2/d +1)}} + \bar{X}_\priv \cdot \frac{n}{n+\nicefrac{m}{(m\tau^2/d +1)}}, 
\end{align}
with $\bar{X}_\pub$ and $\bar{X}_\priv$ being the empirical mean over public and private data points.  Plugging this into \label{eq:breakdown-8}, we get: 
\begin{align}
    \E_{\mu_\priv, v , X} \|[M(X)-\bar{X}]\|_2^2 & \leq \alpha^2 + \frac{
    \nicefrac{md}{\kappa^2}}{(n+m/\kappa)^2} + \frac{nd}{(n+m/\kappa)^2} \nonumber \\
   & \leq \alpha^2 + \frac{
    \nicefrac{md}{\kappa^2}}{(n+m/\kappa)^2} + \frac{nd}{(n+m/\kappa)^2} \nonumber \\
    & = \gO\paren{\alpha^2 + \frac{d}{(n+m/\kappa)}} \label{eq:breakdown-11},
\end{align}
where the second inequality follows from $\kappa \geq 1$. 
Finally, we do the following plugins: we plug in,
\begin{enumerate}
    \item the result from Lemma~\ref{lemma:gaussian-posterior-mean-diff} for the bound on $\E_X\|\E[\mu_\priv \mid X] - \bar{X}\|^2_2$ into \eqref{eq:breakdown-4},
    \item  the result on the bound on $\Ex{\|M(X) - \bar{X}\|_2^2}$ in \eqref{eq:breakdown-11} into \eqref{eq:breakdown-4}, 
    \item the bound on $\Ex{\|\bar{X}-\mu_\priv\|^2_2}$ derived as part of \eqref{eq:breakdown-11} into \eqref{eq:breakdown}
\end{enumerate} 
Together, these complete the proof of the lower bound stated in Lemma~\ref{lemma:gaussian-lb-priv}. 

Finally, we are ready to put together the upper and lower bounds on  $\E[\sum_i Z_i]$ to arrive at the final result in Theorem~\ref{thm:lb-gaussian-dist-shift}. 
From Corollary~\ref{corr:sum-zi-upper-bound-dist-shift} and Lemma~\ref{lemma:gaussian-lb-priv} we conclude that whenever $\tau = \gO(\sqrt{d/m})$, there exists positive constants $c_1, c_2$ and values $n_0, m_0$, such that for $n \geq n_0, m\geq m_0$:
\begin{align*}
   c_2 \cdot d \; \leq \;  \sum_{i=1}^{n} \E[Z_i] + \frac{1}{\kappa} \cdot \sum_{i=n+1}^{n+m} \E[Z_i] & \; \leq   \;c_1\paren{n\varepsilon\alpha +  {{\alpha\sqrt{md}}{}}} 
\end{align*}
This means that when either $n = \Omega(\nicefrac{d}{\varepsilon\alpha})$ or $m = \Omega(\nicefrac{d^2}{\alpha})$. Note that the lower bound in Lemma~\ref{lemma:statistical-lb} is still applicable for $n$. Together, this tells us that when $\tau = \gO(\sqrt{d/m})$, then either $n = \Omega(\nicefrac{d}{\varepsilon\alpha} + \nicefrac{d}{\alpha^2})$ or $m = \Omega(\nicefrac{d}{\alpha^2})$. This completes the first result in Theorem~\ref{thm:lb-gaussian-dist-shift}. For the second part, let us consider the case where $\tau = \omega(\sqrt{d/m})$. Then, either there exists constants $c_1, c_2$, such that for all $n\geq n_0$:
\begin{equation*}
    c_1 d \leq c_2 n\varepsilon \alpha, 
\end{equation*} in which case $n = \Omega(\nicefrac{d}{\varepsilon\alpha})$ or there exists constants $c_1, c_2$ and values $n_0, m_0$, such that for all $n \geq n_0, m \geq m_0$ the following holds:
\begin{equation*}
    d \; \leq \; \frac{\alpha\tau\sqrt{md}+ \alpha \sqrt{md}}{1+m \tau^2/d},
\end{equation*} which implies:
\begin{equation*}
    c_1 d \leq \alpha \tau \nicefrac{\sqrt{md}}{m \tau^2/d}.
\end{equation*} The above holds when $\alpha \gsim \tau$. This completes all conditions in the proof of Theorem~\ref{thm:lb-gaussian-dist-shift}.
\end{proof}

\section{Useful lemmas for linear regression}

We reuse some eigenvalue tail bounds throughout the linear regression lower bounds.

\begin{lemma}[\cite{wainwright2019high}, Theorem 6.1]
\label{lemma:wainwright-eigval-tail}
Let $\mathbf{X} \in \mathbb{R}^{n \times d}$ be drawn according to the $\Sigma$-Gaussian ensemble. 
Then, for $N \geq d$, the minimum singular value $\sigma_{\min}(\mathbf{X})$ satisfies the lower deviation inequality
\[
\mathbb{P} \left[ \frac{\sigma_{\min}(\mathbf{X})}{\sqrt{N}} \leq \lambda_{\min}(\sqrt{\Sigma}) (1 - \delta) - \sqrt{\frac{\mathrm{tr}(\Sigma)}{N}} \right] \leq e^{-N \delta^2 / 2}. 
\]
\end{lemma}

Specializing to the case of $\Sigma = I_d$, we have
\begin{corollary}
\[
\mathbb{P} \left[ \frac{\sigma_{\min}(\mathbf{X})}{\sqrt{N}} \leq 1 - \sqrt{\frac{d}{N}} - \delta \right] \leq e^{-N \delta^2 / 2}. 
\]

and thus,

\[
\mathbb{P} \left[ \sigma_{\min}(\mathbf{X})^2 = \lambda_{\min}(X^\T X) \leq N\left(1 - \sqrt{\frac{d}{N}} - \delta\right)^2 \right] \leq e^{-N \delta^2 / 2}. 
\]
\end{corollary}

We also have the following upper tail variant:
\begin{lemma}[~\cite{wainwright2019high}, Theorem 6.1]
\label{lemma:wainwright-eigval-upper-tail}
Let $\mathbf{X} \in \mathbb{R}^{n \times d}$ be drawn according to the $\Sigma$-Gaussian ensemble. 
Then, for $N \geq d$, the maximum singular value $\sigma_{\max}(\mathbf{X})$ satisfies the upper deviation inequality
\[
\mathbb{P} \left[ \frac{\sigma_{\max}(\mathbf{X})}{\sqrt{N}} \geq \lambda_{\max}(\sqrt{\Sigma}) (1 + \delta) + \sqrt{\frac{\mathrm{tr}(\Sigma)}{N}} \right] \leq e^{-N \delta^2 / 2}. 
\]
\end{lemma}

Specializing to the case of $\Sigma = I_d$, we have
\begin{corollary}
\[
\mathbb{P} \left[ \frac{\sigma_{\max}(\mathbf{X})}{\sqrt{N}} \geq 1 + \sqrt{\frac{d}{N}} + \delta \right] \leq e^{-N \delta^2 / 2}. 
\]

similarly,

\[
\mathbb{P} \left[ \sigma_{\max}(\mathbf{X})^2 = \lambda_{\max}(X^\T X) \geq N\left(1 + \sqrt{\frac{d}{N}} + \delta\right)^2 \right] \leq e^{-N \delta^2 / 2}. 
\]
\end{corollary}

The following simple lemma will be useful in converting exponential tail bounds to upper bounds involving $N$.
\begin{fact}
\label{lem:ex-expansion} For $N \geq 400$ the following is true:
\begin{align} 
e^{-N/8} \leq 1 / N^8.
\end{align}
\end{fact}

The following calculations are also useful in several parts of the proof, so we black-box them here.
\begin{lemma}
\label{lem:t-lowerbound}
Let $z > 100d$ (thus $0 < d < z/100$), $\psi = 1/10$ and consider $t = z(1 - \sqrt{d/z} - \psi)^2$.
Then $t \geq 63/100 \cdot z$.
\end{lemma}
\begin{proof}
We have
\begin{align}
z(1 - \sqrt{d/z} - \psi)^2 &=  z(9/10 - \sqrt{d/z})^2\\
&= z(81/100 - (18/10)\sqrt{d/z} + d/z)\\
&= z(81/100) - (18/10)\sqrt{dz} + d\\
&\geq z(81/100) - (18/10)\sqrt{z^2/100} + d\\
&\geq z(81 - 18)/100 = 63/100 \cdot z.
\end{align}
\end{proof}

\begin{lemma}
\label{lem:t-upperbound}
Let $z > 100d$ (thus $0 < d < z/100$), $\psi = 1/10$ and consider $t = z(1 + \sqrt{d/z} + \psi)^2$.
Then $t \leq 64/100 \cdot z$.
\end{lemma}
\begin{proof}
We have
\begin{align}
z(1 + \sqrt{d/z} + \psi)^2 &=  z(9/10 + \sqrt{d/z})^2\\
&= z(81/100 + (18/10)\sqrt{d/z} + d/z)\\
&= z(81/100) + (18/10)\sqrt{dz} + d\\
&\leq z(81/100) + (18/10)\sqrt{z^2/100} + z/100\\
&\leq z(81 - 18 + 1)/100 = 64/100 \cdot z.
\end{align}
\end{proof}

\newcommand{\bY}{\mathbf{Y}}
\newcommand{\by}{\mathbf{Y}}

\section{Proofs for Linear Regression}

\subsection{Upper bound}

We first upper bound $\Ex{\sum_i Z_i}$.
We can split this into a bound over the sum of public samples 
and private samples.
We will bound the private samples using the differential privacy constraint,
and the public samples using concentration.

For the private samples, we can use Proposition~\ref{prp:prob-to-exp-bound}
as in the mean estimation case.

\begin{proposition}
Assume the conditions in Theorem~\ref{thm:lr-noshift}. Then, $\E[Z_i^\prime] = 0$ and   $\E[(Z_i^\prime)^2] = O(\alpha^2)$, where the expectation is taken over the randomness of sampling $\beta$, $\data$, and the randomness in the mechanism $M$. 
\label{prp:shifted-beta-zi-prime-bound}
\end{proposition}
\begin{proof}
    When conditioned on $\beta$, $(y_i - x_i^\T\beta)x_i$ is independent of $M(\data_i^\prime) - \beta$,  and $\E[(y_i - x_i^\T\beta)x_i \mid \beta] = 0$ (because $\E[\eta_i x_i] = 0$). Thus, we have $\E[Z_i^\prime] = \E[\E[Z_i^\prime \mid \beta]] = 0$.  

    Next, we bound $\E[(Z_i^\prime)^2 | \beta]$ and similarly use the law of total expectation to get the upper bound on $\E[(Z_i^\prime)^2]$.
    Note that we have $y_i = x_i^\T \beta + \eta_i$, thus
    \begin{align*}
        \E[(Z_i^\prime)^2 | \beta] &= \E\brck{\sum_{j,k} (M(\data_i^\prime) - \beta)_j (M(\data_i^\prime) - \beta)_k (\eta_i x_i)_j (\eta_i x_i)_k \vert \beta} \\
        &= \sum_j \E\brck{( M(\data_i^\prime) - \beta)_j^2 (\eta_i x_i)_j^2 \vert \beta}\\
        &= \sum_j \E\brck{( M(\data_i^\prime) - \beta)_j^2 \vert \beta} \Ex{(\eta_i x_i)_j^2 \vert \beta}\\
        &= \Ex{\normtwo{M(\data_i^\prime) - \beta)_j^2}^2 \vert \beta}
        \leq \alpha^2.
    \end{align*}

The second equality uses the fact 
the coordinates of $\eta$ and $x_i$ are sampled independently, 
and the third equality uses 
that $M(\data_i') - \beta$ 
is conditionally independent of $\eta_i x_i$ given $\beta$ 
(because $x_i$ and $y_i$ are resampled in $\data_i'$).
The final step uses the accuracy of the mechanism $M$
and the variance of $\eta_i$ and $x_i$.
\end{proof} 

\begin{proposition}
    Assume the conditions in Theorem~\ref{thm:lr-noshift}. Then, 
    for $i \in [n]$, $\E[Z_i] = O(\eps \alpha)$, where the expectation is taken over the randomness of sampling $\beta$, $\data$, and the randomness in the mechanism $M$.
    \label{prp:shifted-beta-zi-bound}
\end{proposition}
\begin{proof}
    Since $\data$ and $\data_i^\prime$ differ by a single element, and $M$ is $(\eps, \delta)$ differentially private, we can use the fact that the outputs $M(\data)$ and $M(\data_i^\prime)$ (and therefore, by the postprocessing property of DP, $Z_i$ and $Z_i'$) cannot be too different. In particular, we invoke Proposition~\ref{prp:prob-to-exp-bound}. 
\[\left| \mathbb{E}[Z_i] - \mathbb{E}[Z'_i] \right| \leq 2\eps \cdot \mathbb{E}[|Z'_i|] + 2\sqrt{\delta} \cdot \mathbb{E}[Z_i^2 + (Z'_i)^2] \leq 2(\eps + \sqrt{\delta}) \cdot \sqrt{\mathbb{E}[(Z'_i)^2]} + 2\sqrt{\delta \cdot \mathbb{E}[Z_i^2]}.
\]
Next, we apply Proposition~\ref{prp:shifted-beta-zi-prime-bound}.
\[
\mathbb{E}[Z_i] \leq O((\epsilon + \sqrt{\delta})\alpha) + 2\sqrt{\delta \cdot \mathbb{E}[Z_i^2]}.
\]
Next, we bound $\E [Z_i^2]$ with Cauchy-Schwarz followed by the accuracy of $M$, and the concentration of $\eta_i x_i$:
\begin{align*}
    \E [Z_i^2] \;\; &\leq \;\; \sqrt{\E[\|M(\data) - \mu_\priv\|_2^4]}  \sqrt{\E[\|\eta_i x_i\|_2^4]} \\
    &= \;\; O(\alpha^2 d)
\end{align*}
Plugging this result into the equation above, we conclude $\E[Z_i]=O(\epsilon \alpha + \sqrt{\delta d}\alpha)$. Recall from the conditions in Theorem~\ref{thm:lr-noshift}, $\delta \leq \nicefrac{\epsilon^2}{d}$. Thus, $\E[Z_i] = O(\epsilon \alpha)$.
\end{proof}

Finally, we bound the value of $Z_i$ for the public samples.

\begin{proposition}
    \label{prp:upper-bound-zi-public} Assume the conditions in Theorem~\ref{thm:lb-gaussian-no-dist-shift}.      Then, the following holds for the expected sum of the fingerprinting statistics defined for the public samples: $\E[\sum_{i=n}^{n+m}Z_i] =\gO(\alpha\sqrt{md})$.
\end{proposition}
\begin{proof}
Recall the definition of the fingerprinting statistics: $\E[Z_j] = \E\left[\inprod{M(x) - \beta}{(y_i - x_i^\T \beta)x_i}\right]$. Then,
\begin{align}
    \E\left[\sum_{i=n}^{n+m} Z_i\right] &= \E\left[\inprod{M(\data) - \beta_\priv}{\sum_{i=n}^{n+m}(y_i - x_i^\T \beta)x_i}\right]\\
    &\leq \sqrt{\E[\norm{M(\data)-\beta}{2}^2]}\sqrt{\sum_{i=n}^{n+m}\E[ \norm{\eta_ix_i}{2}^2]} \\
    &= \gO(\alpha \sqrt{md}).
\end{align}
The first inequality uses Cauchy-Schwarz, and the second uses the bound on the accuracy of the mechanism $M$ and the variance of $\eta_ix_i$.  
\end{proof}
\begin{corollary}
    \label{corr:sum-zi-upper-bound}
    Under the conditions of Theorem~\ref{thm:lr-noshift}, for the expectation taken over the randomness of $\beta, X_\priv, X_\pub, \eta$ and the mechanism $M$, $\sum_{i=1}^{n+m} \E[Z_i] = O(n\varepsilon\alpha + \alpha\sqrt{md})$.
\end{corollary}

\subsection{Fingerprinting lemma (lower bound)}

\begin{theorem}
\label{thm:lr-noshift-fingerprinting}
Assume the conditions of Theorem~\ref{thm:lr-noshift}.
Then 
\begin{align}
\Ex{\sum_i Z_i} = \Omega(d)
\end{align}
where the sum is taken over both the public and private samples.
\end{theorem}

To prove this theorem, 
we follow the same structure as in the mean estimation proof
(conditioning on $X, y$ instead of just $X$).

Let $\hat{\beta} \eqdef (X^TX + \lambda {I}_d)^{-1}X^T y$ for some $\lambda > 0$. Then, we again have the breakdown:
\begin{align}
\Ex{\sum_i Z_i} &= \Ex{ \inprod{M(X, y) - \beta}{X^T y - X^TX \beta}}\\
&= \Ex{ \inprod{M(X, y) - \hat{\beta} + \hat{\beta} - \beta}{X^TX (\hat{\beta} - \beta)}}\\
&= \Ex {\|X\beta  - X\hat{\beta}\|_2^2} - \Ex{ \inprod{M(X, y) - \hat{\beta} }{X^TX (\beta - \hat{\beta} )}}\\
&= \Ex{\normtwo{X(\hat{\beta} - \beta)}^2} - \Ex{\inprod{M(X, y) - \hat{\beta}}{X^T(X\beta) - X^\T y}}
\end{align}

The first term corresponds to the expected error of the nonprivate OLS estimate. We include a short proof for completeness.
\begin{lemma} Under the assumptions of Theorem~\ref{thm:lr-noshift},
\[
\varEx{\beta, X, \eta}{\normtwo{X(\hat{\beta} - \beta)}^2} = \sigma^2d
\]
\end{lemma}
\begin{proof}
Rewrite $\hat{\beta} = (X^\T X)^{-1}X^\T y = (X^\T X)^{-1}X^\T (X\beta + \eta) = \beta + (X^\T X)^{-1}X^\T \eta$. 

Then $\hat{\beta} - \beta = (X^\T X)^{-1}X^\T \eta$. Hence, $X(\hat{\beta} - \beta) = X(X^\T X)^{-1}X^\T \eta$.

Finally 
\begin{align}
\Ex{\normtwo{X(X^\T X)^{-1}X^\T \eta}^2} &= \Ex{\Tr(\eta^T X(X^\T X)^{-1}X^\T \eta)}\\
&= \sigma^2 d
\end{align}

\end{proof}

We now proceed similarly to the mean estimation proof in order to upper bound the second term. Conditioning on $X, \eta$ we have:

\begin{align}
&\varCondEx{M, \beta}{\inprod{M(\cal{D}) - \hat{\beta}}{X^\T X \beta - X^\T y}}{X, y} \\
&= {\inprod{\E_{M}\brck{M(\cal{D}) - \hat{\beta} \;|\; X, y}}{ \E_\beta \brck{ X^\T X \beta - X^\T y \;|\; X, y}}}\\
&= \normtwo{\varCondEx{M}{M(\cal{D}) - \hat{\beta}}{X, y}} \cdot \normtwo{\varCondEx{\beta}{X^\T X \beta - X^\T y}{X, y}}\\
&\leq \sqrt{\varCondEx{M}{\normtwo{M(\cal{D}) - \hat{\beta}}^2}{X, y}} \cdot \normtwo{\varCondEx{\beta}{X^\T X \beta - X^\T y}{X, y}}
\end{align}

Removing the conditioning on $X, y$ and applying Cauchy-Schwarz, we have

\begin{align}
\sqrt{
    \mathbb{E}_{X,y,M} \normtwo{M(X,y) - \hat{\beta}}^2 
    \cdot
    \mathbb{E}_{X, y}\normtwo{X^\top X \mathbb{E}_{\beta}[\beta|X,y]  - X^\top y}^2
}
\end{align}
\begin{tikzpicture}[overlay]
    \node at (5.6,0) {
    \textcolor{red}{
    $\underbrace{\hspace{3.6cm}}_{
    \textcolor{red}{
    \tikz[baseline]{\node[circle,draw=red,inner sep=0pt] (1) {1};}}}$}};
    \node at (10.4,0) {
    \textcolor{blue}{$\underbrace{\hspace{5cm}}_{
    \textcolor{blue}{
    \tikz[baseline]{\node[circle,draw=blue,inner sep=0pt] (2) {2};}}}$}};
\end{tikzpicture}

\bigskip

To bound term \textcolor{red}{\circled{1}}, 
we use the triangle inequality as before:
\begin{align}
   \mathbb{E}_{X,\eta} \mathbb{E}_{\hat{\beta}} \left\| M(X,y) - \hat{\beta} \right\|_2^2 &= \mathbb{E}_{X,\eta} \mathbb{E}_{\hat{\beta}} \left\| M(X,y) - \beta_\priv + \beta_\priv - \hat{\beta} \right\|_2^2   \\
   &\leq \alpha^2 +  \mathbb{E}_{X,\eta} \mathbb{E}_{\hat{\beta}} \left\| \hat{\beta} - \hat{\beta} \right\|_2^2  \nonumber \\
   & \leq \alpha^2 + \sigma^2 \cdot \nicefrac{d}{n}.
\end{align}

We will next bound the term \textcolor{blue}{\circled{2}}.

We need the posterior mean $\Ex{\beta \vert X, y}$.
We choose the prior $\beta \sim \cal{N}(0, b^{-1}I_d)$ where $b$ is the precision.
This gives the posterior $\beta \vert X, y \sim \cal{N}(\mu, \Lambda^{-1})$
where
\begin{align}
\Lambda = aX^\T X + bI, \quad \mu = a\Lambda^{-1}X^\T y.
\end{align}
The above is a standard derivation for computing the posterior of the linear regression parameter under a Bayesian prior (see Chapter 11.7 in \citet{pml1Book} for reference). The parameter $a$ here is the noise variance $\sigma^2$.

Then we have,
\begin{align}
\E_{\beta}[\beta \vert X, y] = a\Lambda^{-1}X^\T y
\end{align}

Now to remove the conditioning, we can expand
\begin{align}
\mathbb{E}_{X,y} \left\| X^\top X\mathbb{E}_{\beta}[\beta|X,y]  - X^\top y \right\|_2^2 &= \Ex{\normtwo{X^\T X\Ex{\beta \vert X, y} - X^\T y}^2}\\
&= \Ex{\normtwo{X^\T X a\Lambda^{-1}X^\T y - X^\T y}^2}\\
&= \Ex{\normtwo{(X^\T X a\Lambda^{-1} - I_d)X^\T y}^2}
\end{align}

Then 
\begin{align}
&= \Ex{\normtwo{(X^\T X a\Lambda^{-1} - I_d)X^\T y}^2}\\
&\leq \sqrt{\Ex{\normtwo{(X^\T X a\Lambda^{-1} - I_d)}^4}\Ex{\normtwo{X^\T y}^4}}\\
&= \sqrt{\Ex{\normtwo{(X^\T X a\Lambda^{-1} - \Lambda\Lambda^{-1})}^4}\Ex{\normtwo{X^\T y}^4}}\\
&\leq \sqrt{\Ex{\normtwo{(X^\T X a - \Lambda)}^4\normtwo{\Lambda^{-1}}^4}\Ex{\normtwo{X^\T y}^4}}\\
&= \sqrt{\Ex{\normtwo{(X^\T X a - (X^\T Xa + bI))}^4\normtwo{\Lambda^{-1}}^4}\Ex{\normtwo{X^\T y}^4}}\\
&= \sqrt{b^4\Ex{\normtwo{\Lambda^{-1}}^4}\Ex{\normtwo{X^\T y}^4}}
\end{align}

We now derive upper bounds on $\Ex{\normtwo{\Lambda^{-1}}^4}$ and $\Ex{\normtwo{X^\T y}^4}$.

\begin{lemma}
Under the conditions of Theorem \ref{thm:lr-noshift}, $\Ex{\normtwo{\Lambda^{-1}}^4} \leq \gO\paren{\frac{1}{(aN+b)^4}}$, when $N = m+n = \Omega(d)$ and $b > 1/N$. 
\label{lem:first-term-upper-bound}
\end{lemma}

\begin{proof}
First, note that we can rewrite $\Ex{\normtwo{\Lambda^{-1}}^4}$ as follows:
\begin{align}
\Ex{\normtwo{\Lambda^{-1}}^4} &= \Ex{\normtwo{(aX^\T X + bI)^{-1}}^4}\\
&= \Ex{\frac{1}{(a\sigma_{\min}(X^\T X) + b)^4}}
\end{align}

We will split the expectation into two conditional expectations:
\begin{align}
\Ex{\normtwo{\Lambda^{-1}}^4} = \condEx{\normtwo{\Lambda^{-1}}^4}{\sigma_{\min}(X^\T X) \in (0, t]}\Pr[\sigma_{\min}(X^\T X) \in (0, t)] + \\
\condEx{\normtwo{\Lambda^{-1}}^4}{\sigma_{\min}(X^\T X) \in [t, \infty)]}\Pr[\sigma_{\min}(X^\T X) \in [t, \infty)].
\end{align}
for some $0 < t < 1$.

For any $\sigma \geq 0$ we can bound $\Ex{\normtwo{\Lambda^{-1}}^4} < 1/b^4$,
and when $\sigma \geq t$ we additionally have
$\Ex{\normtwo{\Lambda^{-1}}^4} \leq 1/(at + b)^4$.
Then the above simplifies to
\begin{align}
&\condEx{\normtwo{\Lambda^{-1}}^4}{\sigma_{\min}(X^\T X) \in (0, t]}\Pr[\sigma_{\min}(X^\T X) < t] + \\
&\condEx{\normtwo{\Lambda^{-1}}^4}{\sigma_{\min}(X^\T X) \in [t, \infty)]}\Pr[\sigma_{\min}(X^\T X) \in [t, \infty)]\\
&= \frac{1}{b^4}\Pr[\sigma_{\min}(X^\T X) < t] + \frac{1}{(at+b)^4}\Pr[\sigma_{\min}(X^\T X) \geq t]
\end{align}
We approximate $\Pr[\sigma_{\min}(X^\T X) \geq t] \leq 1$, giving
\begin{align}
&= \frac{1}{b^4}\Pr[\sigma_{\min}(X^\T X) < t] + \frac{1}{(at+b)^4}
\end{align}

It then remains to bound $\Pr[\sigma_{\min}(X^\T X) < t]$.

As $X^\T X$ is symmetric and PSD, we have $\sigma_{\min}(X^\T X) = \lambda_{\min}(X^\T X) = \sigma_{\min}(X)^2$.

Now we have
\begin{align}
\Pr[\sigma_{\min}(X^\T X) \in (0, t]] &\leq \Pr(\sigma_{\min}(X^\T X) < t)\\
&= \Pr(\sigma_{\min}(X)^2 < t)
\end{align}

We can now use Lemma~\ref{lemma:wainwright-eigval-tail}.
Setting $t = N(1-\sqrt{d/N} - \psi)^2$ for some $0 < \psi < 1 - \sqrt{d/N}$, $\Pr(\sigma_{\min}(X)^2 < t) \leq e^{-N\psi^2/2}$.

Using Fact~\ref{lem:ex-expansion}, we can upper bound this by, e.g., $O(1/N^8)$.

Thus, we have $\frac{1}{b^4}\Pr[\sigma_{\min}(X^\T X) < t] = O(\frac{1}{b^4N^8})$, so in total we have
\begin{align}
\Ex{\normtwo{\Lambda^{-1}}^4} &\leq \frac{1}{b^4}\Pr[\sigma_{\min}(X^\T X) < t] + \frac{1}{(at+b)^4}\\
&\leq O\left(\frac{1}{b^4N^8}\right) + O\left(\frac{1}{(at+b)^4}\right)
\end{align}

Assuming $N > 100d$, we have the constraint $0 < \psi < 9/10$. 
We set $\psi = 1/10$.
Then this gives 
\begin{align}
t &= N(1-\sqrt{d/N} - 1/10)^2 \\
&= N(9/10)^2 - 2(9/10)\sqrt{dN} + d
\end{align}

Since we assume $N > 100d$ (i.e. $d < N/100$), 
\begin{align}
&\geq  N(81/100) - (18/10)\sqrt{N^2/100} + d\\
&= N(81-18)/100 + d = \Omega(N).\label{eq:eigval-const}
\end{align}

Thus, for $b > 1/N$, we have
\begin{align}
O\left(\frac{1}{b^4N^8}\right) + O\left(\frac{1}{(at+b)^4}\right) \leq O\left(\frac{1}{b^4N^8}\right) + O\left(\frac{1}{(aN+b)^4}\right) \leq O\left(\frac{1}{(aN+b)^4}\right).
\end{align}
\end{proof}

\begin{lemma}
    Under the conditions of Theorem ~\ref{thm:lr-noshift}, $\Ex{\|X^T y\|_2^4} = \mathcal{O}\paren{N^2 d^2 + N^4 d^2 / b^2}$, where $N=m+n$.  
    \label{lem:second-term-upper-bound}
\end{lemma}
\begin{proof}
    Let $N = m+n$ be the total number of datapoints (including both public and private). First, we simplify $X^T y = X^T(X\beta + \Vec{\eta})$.
    Then, we use the identity $(a+b)^2 \leq 2(a^2 +b^2)$ twice and get:
    \begin{align}
        \label{eq:100}
        \Ex{\|X^T y\|_2^4} \leq \Ex{\|X^TX \beta\|_2^4} + \Ex{\|X^T\Vec{\eta}\|_2^4}.
    \end{align}
    Now, we bound each of the above two terms separately. First, we start with the second term $\Ex{\|X^T\eta\|_2^4}$. 
    \begin{align*}
        \Ex{\|X^T\Vec{\eta}\|_2^4} &= \Ex{\normtwo{\sum\nolimits_{i\in [N]} x_i \eta_i}^4} \\
        &= \Ex{{\sum\nolimits_{i,j,k,l\in [N]} x_i^T x_j x_k^T x_l \cdot \eta_i \eta_j \eta_k \eta_l}}
    \end{align*}
Since $\eta_i \perp \eta_j$ whenever $i \neq j$ and $\Ex{\eta} = 0$, it is easy to see that the only non-zero terms in the expectation above are when: (i) $i=j=k=l$; (ii) $i=j, k=l, i \neq k$; (iii) $(i,j)=(k,l), i \neq j$. Thus, the summation breaks down into:
\begin{align}
    \label{eq:101}
    \Ex{\|X^T\Vec{\eta}\|_2^4} &= N \Ex{\normtwo{x}^4} \Ex{\eta^4} + N (N-1) \paren{\Ex{\normtwo{x}^2} \Ex{\eta^2}}^2 \nonumber \\
    & \quad +  N (N-1) \Ex{\normtwo{x}^2} \paren{\Ex{\eta^2}}^2  \\
    &= 3N\Ex{\normtwo{x}^4} + N(N-1) \paren{\Ex{\normtwo{x}^2}}^2 + \Ex{\normtwo{x}^2} \nonumber \\
    &= \gO(N^2 d^2). \nonumber
\end{align}
In the above equations, the first equality uses  $x_i \perp x_j $ when $i \neq j$, and $x_i \perp \eta_i, \forall i$. The second uses the fact that $\Ex{\eta^4} = 3 $ and $\Ex{\eta^2} = 1$. Finally, the last equality uses $\Ex{\normtwo{x}^2} = d$ and $\Ex{\normtwo{x}^4} = \gO(d^2)$. 

Next, we look at the first term in \eqref{eq:100}.
Using the independence of $X$ and $\beta$, we bound this as follows:
\begin{align}
\Ex{\normtwo{X^\T X\beta}^4} &\leq \Ex{\normtwo{X^\T X}^4 \normtwo{\beta}^4}\\
&= \Ex{\normtwo{X^\T X}^4}\Ex{\normtwo{\beta}^4}
\end{align}

Now we have $\Ex{\normtwo{\beta}^4} = O(\frac{d^2}{b^2})$.
Further, we can bound $\Ex{\normtwo{X^\T X}^4} = O(N^4)$.

This gives an overall bound of $\Ex{\normtwo{X^\T X\beta}^4} = O(N^4d^2/b^2)$. Combining with the previous term gives the result.

\end{proof}     

\begin{lemma}
    When $b=1/d > 1/N$, $m+n = \Omega(d)$, and $\alpha = \gO(1)$, the expression: 
    \begin{align*}    
    \sqrt{
    \mathbb{E}_{X,\eta} \mathbb{E}_{\hat{\beta}} \left\| M(X,y) - \hat{\beta} \right\|_2^2 
    \cdot
    \mathbb{E}_{X,\eta} \left\| X^\top X \mathbb{E}[\beta|X,\eta]  - X^\top y \right\|_2^2
} = \gO(1).
\label{lem:lb-nodist-shift-lr-zi-lb}
\end{align*}
\end{lemma}

\begin{proof}
    From Lemma~\ref{lem:first-term-upper-bound} and Lemma~\ref{lem:second-term-upper-bound}, we derive the following:
    \begin{align}
       & \sqrt{\mathbb{E}_{X,\eta} \mathbb{E}_{\hat{\beta}} \left\| M(X,y) - \hat{\beta} \right\|_2^2 
    \cdot \sqrt{b^4\Ex{\normtwo{\Lambda^{-1}}^4}\Ex{\normtwo{X^\T y}^4}}}\\
   & \leq \sqrt{\mathbb{E}_{X,\eta} \mathbb{E}_{\hat{\beta}} \left\| M(X,y) - \hat{\beta} \right\|_2^2 \cdot b^2 (\nicefrac{1}{(aN+b)^2}) \cdot \sqrt{N^4d^2/b^2 + N^2d^2} } \nonumber  \\
   &\leq \sqrt{\left(\sigma^2d/N + \alpha^2 \right) b^2 (\nicefrac{1}{(aN+b)^2}) \cdot (\sqrt{N^4d^2/b^2} + \sqrt{N^2d^2})}\\
   &\leq \sqrt{\left(\sigma^2d/N + \alpha^2 \right) b^2 (\nicefrac{1}{(aN)^2}) \cdot (N^2d/b + Nd)} \\
    &\leq \sqrt{\left(\sigma^2d/N + \alpha^2 \right) b \cdot \left(\frac{d}{a^2} + \frac{db}{a^2 N}\right)}  
   \end{align}
Note that generally we will set $a = \sigma^2$.
Let $b = 1/d$. Also, note $\alpha = O(1)$. Then:
\begin{align}
    &\leq \sqrt{\left(\sigma^2d/N + \alpha^2 \right) (1/d) \cdot (\sigma^2d + \sigma^4/N)}\\
        &\leq \sqrt{\left(\sigma^2d/N + \alpha^2 \right) \cdot (\sigma^4 + \sigma^4/(Nd))}\\
        &= O(1),
\end{align}
since $\alpha^2 = \gO(1)$. \end{proof}

Now, we are ready to prove the final result in Theorem~\ref{thm:lb-linreg-dist-shift-informal}. For this we invoke the upper bound on $\Ex{\sum_i Z_i}$ in Corollary~\ref{corr:sum-zi-upper-bound} and the lower bound of $\Ex{\sum_i Z_i} = \Omega(d)$ induced by \eqref{eq:main-breakdown}, Lemma~\ref{lemma:breakdown} and Lemma~\ref{lem:lb-nodist-shift-lr-zi-lb}. This tells us that there exist positive constants $c_1, c_2$ and values $m_0, n_0$ such that for $m\ge m_0, n\ge n_0$, the following holds:
\begin{align}
 c_1 \cdot d   \;\le \;  \Ex{\sum_{i\in[N]} Z_i} \; \le  
    \; c_2 \cdot  \paren{n\varepsilon\alpha + \alpha\sqrt{md}}
\end{align}
Thus, either $m = \Omega(\frac{d}{\alpha^2})$ or $n \geq \Omega(\frac{d}{\varepsilon \alpha})$. Note that, similar to Lemma~\ref{lemma:statistical-lb} for Gaussian mean estimation, a similar statistical lower bound also applies for linear regression which implies $n =\Omega(\frac{d}{\alpha^2})$. Together this tells us, that when public and private data come from the same distribution, then for privately solving linear regression with an $\varepsilon, \delta$-DP mechanism, where $\delta \le \varepsilon^2/d$, to a parameter estimation accuracy of $\alpha$ in $\ell_2$, we either need $m = \Omega(d/\alpha^2)$ public samples or the number of public and private samples together must satisfy: $n+m = \Omega(\tfrac{d}{\alpha\varepsilon}+\tfrac{d}{\alpha^2})$. This completes the proof of our result in Theorem~\ref{thm:lr-noshift}.

\section{Linear Regression with Shifted Parameter}
\label{sec:lr-shifted}

We now consider a variation of linear regression where the public parameter $\beta_{\pub}$
is shifted by a random vector $v$ sampled from $\cal{N}(\mathbf{0}, \tau^2/d \cdot I_d)$.
The goal is to use both the public and private samples to estimate the private parameter $\beta_{\priv}$.

That is, we have $n$ samples $(x_i, y_i)$ such that
\begin{align}
y_i = x_i^\T \beta_{\priv} + \eta_i
\end{align}

and $m$ samples such that
\begin{align}
y_i &= x_i^\T \beta_{\pub} + \eta_i\\
&= x_i^\T (\beta_{\priv} + v) + \eta_i
\end{align}

Let $X^\T = [X_\pub^\T \; \; X_\priv^\T]$.
Note that $y = X\beta + Pv + \eta$ where 
\begin{align}
P &= \begin{bmatrix}
X_\pub\\
0_{n \times d}
\end{bmatrix}
\end{align}

We can define $\Psi = Pv + \eta$ to be noise correlated with $X_\pub$.

Set $\hat{\beta}$ to be the generalized least squares (GLS) estimator~\citep{aitken1936iv} $(X^\T \Sigma^{-1}X)^{-1}X^\T \Sigma^{-1} y$,
where $\Sigma = \Cov(\Psi | X)$. 
In particular because $P, v, \eta$ are independent and mean zero,
\begin{align}
\Sigma &= \Ex{(Pv + \eta)(Pv + \eta)^\T | X} = \Ex{Pvv^\T P^\T} + \Ex{\eta \eta^T}\\
&= (\tau^2/d)PP^\T + \sigma^2 I_N
\end{align}

Then we have $(X^\T \Sigma^{-1}X)\hat{\beta} = X^\T \Sigma^{-1}y$.

With this in mind, we will use a modified test statistic:
\begin{align}
\Ex{\sum_i Z_i} = \Ex{\inprod{M(X, y) - \beta}{X^\T \Sigma^{-1}(X\beta - y)}}
\end{align}

\subsection{Upper bound}

As in the non-shifted proof, for the upper bound, 
we split the sum over fingerprinting statistics into public and private samples.
For the private samples we will use the differential privacy constraint to bound the sum over samples (as in the non-shifted case),
while for the public samples we can use concentration while taking 
into account the parameter shift.

Note that $\Sigma$ is block-diagonal,
so that 
\begin{align}
\Sigma^{-1} &= \begin{bmatrix}
(\frac{\tau^2}{d}X_\pub X_\pub^\T + \sigma^2 I_m)^{-1} & 0\\
0 & 1/\sigma^2 I_n
\end{bmatrix}
\end{align}

Hence, $X^\T \Sigma^{-1} = [X_\pub^T(\frac{\tau^2}{d}X_\pub X_\pub^\T + \sigma^2 I_m)^{-1} \;\; \frac{1}{\sigma^2}X_\priv^\T]$.
This implies that we can analyze the sum over private samples
using the same analysis as in the non-shifted setting,
using ~\ref{prp:shifted-beta-zi-bound},
up to a factor of $1/\sigma^2$.

It then remains to bound the sum over public samples using concentration.

\begin{lemma}
\label{lem:shifted-lr-ub}
$\Ex{\sum_{i=n}^{n+m}Z_i} \leq O\left(n\eps \alpha + \alpha\sqrt{\frac{1}{\sigma^2}md - \frac{1}{\sigma^4}\left(\frac{m^2d}{\frac{d}{\tau^2} + \frac{1}{\sigma^2}m} \right)}\right)$.
\end{lemma}

\begin{proof}
Let $\Sigma^{-1}_{\pub} = (\Sigma^{-1})_{11} = (\frac{\tau^2}{d}X_\pub X_\pub^\T + \sigma^2 I_m)^{-1}$. Then we have
\begin{align}
\Ex{\sum_{i=n}^{n+m} Z_i} &= \Ex{\inprod{M(X, y) - \beta}{X_{\pub}^\T \Sigma^{-1}_{\pub}(X_\pub\beta - y_\pub)}}\\
&= \Ex{\inprod{M(X, y) - \beta}{X_{\pub}^\T \Sigma^{-1}_{\pub}(X_\pub\beta - (X_\pub\beta + X_\pub v + \eta_\pub)}}\\
&\leq \sqrt{\Ex{\normtwo{M(X, y) - \beta}^2}}\sqrt{\Ex{\normtwo{X_\pub^\T\Sigma^{-1}_\pub(X_\pub v + \eta_\pub)}^2}}
\end{align}

We first bound just the second term.
\begin{align}
&\Ex{\normtwo{X_\pub^\T\Sigma^{-1}_\pub(X_\pub v + \eta_\pub)}^2} \\
&= \Ex{\Tr\left((X_\pub v + \eta_\pub)^\T \Sigma^{-1}_\pub X_\pub X_\pub^\T\Sigma^{-1}_\pub(X_\pub v + \eta_\pub)\right)}\\
&= \Ex{\Tr\left(\Sigma^{-1}_\pub X_\pub X_\pub^\T\Sigma^{-1}_\pub(X_\pub v + \eta_\pub)(X_\pub v + \eta_\pub)^\T \right)}\\
&= \Tr(\varEx{X}{\Ex{\Sigma^{-1}_\pub X_\pub X_\pub^\T\Sigma^{-1}_\pub \vert X}\Ex{(X_\pub v + \eta_\pub)(X_\pub v + \eta_\pub)^\T\vert X}})\\
&= \Tr(\varEx{X}{\Sigma^{-1}_\pub X_\pub X_\pub^\T\Sigma^{-1}_\pub\Sigma_\pub})\\
&= \varEx{X}{\Tr(X_\pub^\T\Sigma^{-1}_\pub X_\pub)}
\end{align}

Applying the Woodbury identity to $\Sigma^{-1}_\pub$, we get
\begin{align}
&= \varEx{X}{\Tr(X_\pub^\T (\frac{1}{\sigma^2}I_m - \frac{1}{\sigma^4}X_\pub^\T(\frac{d}{\tau^2}I_d + \frac{1}{\sigma^2}X_\pub^\T X_\pub)^{-1}X_\pub^\T)X)}\\
&= \varEx{X}{\Tr(X_\pub^\T\frac{1}{\sigma^2}X)} - \varEx{X}{\Tr(X_\pub^\T(\frac{1}{\sigma^4}X_\pub(\frac{d}{\tau^2}I_d + \frac{1}{\sigma^2}X_\pub^\T X_\pub)^{-1}X_\pub^\T)X)}
\end{align}

The first term is $\frac{1}{\sigma^2}md$ by standard Frobenius norm bounds. Then we need to lower bound the second term:
\begin{align}
&\varEx{X}{\Tr(\frac{1}{\sigma^4}(\frac{d}{\tau^2}I_d + \frac{1}{\sigma^2}X_\pub^\T X_\pub)^{-1}X_\pub^\T X_\pub X_\pub^\T X)}\\
&\geq \varEx{X}{\frac{1}{\sigma^4}\normtwo{(\frac{d}{\tau^2}I_d + \frac{1}{\sigma^2}X_\pub^\T X_\pub)^{-1}X_\pub^\T X_\pub X_\pub^\T X}}\\
&\geq \frac{1}{\sigma^4}\varEx{X}{\lambda_{\min}(\frac{d}{\tau^2}I_d + \frac{1}{\sigma^2}X_\pub^\T X_\pub)^{-1}\Tr(X_\pub^\T X_\pub X_\pub^\T X_\pub)}\\
&= \frac{1}{\sigma^4}\varEx{X}{\frac{1}{\normtwo{(\frac{d}{\tau^2}I_d + \frac{1}{\sigma^2}X_\pub^\T X_\pub)}}\normf{X_\pub^\T X_\pub}^2}\\
&\geq \frac{1}{\sigma^4}\varEx{X}{\frac{1}{\normtwo{(\frac{d}{\tau^2}I_d + \frac{1}{\sigma^2}X_\pub^\T X_\pub)}}\left(\sum_{j=1}^d \lambda_j(X_\pub^\T X_\pub)^2\right)}\\
&\geq \frac{1}{\sigma^4}\varEx{X}{\frac{1}{\frac{d}{\tau^2} + \frac{1}{\sigma^2}\lambda_{\max}(X_\pub^\T X_\pub)}\left(d\lambda_{\min}(X_\pub^\T X_\pub)^2\right)}
\end{align}

We can break the above conditional expectation into two components, one where $\lambda_{\min}(X_\pub^\top X_\pub) \in (0, t_1], \lambda_{\max}(X_\pub^\top X_\pub) \in [t_2, \infty), t_1 \leq t_2$, and the other given by its complement, for some appropriately chosen $t_1, t_2> 0$. 

\begin{align*}
    &\frac{1}{\sigma^4}\varEx{X}{\frac{1}{\frac{d}{\tau^2} + \frac{1}{\sigma^2}\lambda_{\max}(X_\pub^\T X_\pub)}\left(d\lambda_{\min}(X_\pub^\T X_\pub)^2\right)} \\
    &\geq  \frac{1}{\sigma^4} \cdot \frac{dt_1^2}{\nicefrac{d}{\tau^2} + \nicefrac{t_2}{\sigma^2}} \Pr\paren{\lambda_{\min}(X_\pub^\top X_\pub) \geq t_1} \cdot \Pr\paren{\lambda_{\max}(X_\pub^\top X_\pub) \leq t_2}
\end{align*}

The two tail bounds on the eigen values of $X_\pub^\top X_\pub$ that appear in the above equation can be resolved using Lemma~\ref{lemma:wainwright-eigval-tail}, and Lemma~\ref{lemma:wainwright-eigval-upper-tail}. When $N \geq 100d$, then with constant probability $\lambda_{\min} (X_\pub^\top X_\pub) = m - \Omega(1)$ and $\lambda_{\max} (X_\pub^\top X_\pub) = m + \gO(1).$ Plugging this result into the above lower bound, we conclude:
\begin{align}
\frac{1}{\sigma^4}\varEx{X}{\frac{1}{\frac{d}{\tau^2} + \frac{1}{\sigma^2}\lambda_{\max}(X_\pub^\T X_\pub)}\left(d\lambda_{\min}(X_{\pub}^\T X_{\pub})^2\right)} \geq
\frac{1}{\sigma^4}\Omega\left(\frac{dm^2}{\frac{d}{\tau^2} + \frac{1}{\sigma^2}m}\right)
\end{align}

Thus, in total, we have: 
\begin{align}
\Ex{\normtwo{X_\pub^\T\Sigma^{-1}_\pub(X_\pub v + \eta_\pub)}^2} \leq \frac{1}{\sigma^2}md - \frac{1}{\sigma^4}\left(\frac{m^2d}{\frac{d}{\tau^2} + \frac{1}{\sigma^2}m} \right).
\end{align}

Finally, from the accuracy assumption of the mechanism $M$ we have $\sqrt{\Ex{\normtwo{M(X, y) - \beta}^2}} \leq \alpha$. Putting it all together, we arrive at the inequality:
\begin{align}
\Ex{\sum_{i=n}^{n+m} Z_i}  = \gO \paren{ \alpha\sqrt{\frac{1}{\sigma^2}md - \frac{1}{\sigma^4}\left(\frac{m^2d}{\frac{d}{\tau^2} + \frac{1}{\sigma^2}m} \right)}}.
\end{align}

The statement of Lemma~\ref{lem:shifted-lr-ub} follows from applying the non-shifted upper bound over private samples in Proposition ~\ref{prp:shifted-beta-zi-bound} along with the above result.

\end{proof}

\subsection{Lower bound}

Transforming the test statistic appropriately gives:
\begin{align}
\Ex{\sum_i Z_i} = \Ex{\normtwo{\Sigma^{-1/2}X(\hat{\beta} - \beta)}^2} - \Ex{\inprod{M(X, y) - \hat{\beta}}{X^\T \Sigma^{-1}(X\beta - y)}} \label{eq:1001}
\end{align}

In the above, we lower bound the first term with the non-private statistical error. For the second term,   since $\Sigma$ only depends on $X_\pub$ and $\sigma^2$,
we can still break the dependence in the terms as usual by conditioning on $X$ and $y$.
\begin{align}
&\E_{X,y}\brck{\varCondEx{M, \beta, v}{\inprod{M(X, y) - \hat{\beta}}{X^\T \Sigma^{-1} X \beta - X^\T \Sigma^{-1} y}}{X, y}} \\
&= \E_{X,y}\brck{{\inprod{\E_{M}[{M(X, y) - \hat{\beta} \;|\; X, y}]}{ \E_{\beta, v} \brck{ X^\T\Sigma^{-1} X \beta - X^\T \Sigma^{-1} y \;|\; X, y}}}}\\
&\leq \sqrt{\E_{X,y}\E_M{\normtwo{M(X, y) - \hat{\beta}}^2}} \cdot \sqrt{\E_{X, y}\normtwo{\varCondEx{\beta, v}{X^\T \Sigma^{-1} X \beta - X^\T \Sigma^{-1} y}{X, y}}^2}
\end{align}

The final inequality uses Cauchy-Schwarz. Removing the conditioning on $X, y$ we get:

\begin{align}
\sqrt{
    \mathbb{E}_{X,y,M} \normtwo{M(X,y) - \hat{\beta}}^2 
    \cdot
    \mathbb{E}_{X, y}\normtwo{X^\top \Sigma^{-1} X \cdot  \mathbb{E}_{\beta, v}[\beta|X,y]  - X^\top \Sigma^{-1} y}^2
}\label{eq:1002}
\end{align}
\begin{tikzpicture}[overlay]
    \node at (4.7,0) {
    \textcolor{red}{
    $\underbrace{\hspace{3.6cm}}_{
    \textcolor{red}{
    \tikz[baseline]{\node[circle,draw=red,inner sep=0pt] (1) {1};}}}$}};
    \node at (10.6,0) {
    \textcolor{blue}{$\underbrace{\hspace{7.cm}}_{
    \textcolor{blue}{
    \tikz[baseline]{\node[circle,draw=blue,inner sep=0pt] (2) {2};}}}$}};
\end{tikzpicture}
\vspace{0.5em}

Next, we present a lemma that lower bounds the first term in \eqref{eq:1001}.

\begin{lemma}
\label{lem:shifted-lr-statistical-error}
$\Ex{\normtwo{\Sigma^{-1/2}X(\hat{\beta}-\beta)}^2} = \Theta(d)$.
\end{lemma}

\begin{proof}
First, we simplify $\hat{\beta} - \beta$:
\begin{align}
\hat\beta - \beta &= (X^\T \Sigma^{-1}X)^{-1}X^\T \Sigma^{-1}(X\beta + Pv + \eta) - \beta\\
&= (X^\T \Sigma^{-1}X)^{-1}X^\T \Sigma^{-1}(Pv + \eta)
\end{align}

Then we have,
\begin{align}
\Ex{\normtwo{\Sigma^{-1/2}X(\hat{\beta}-\beta)}^2} &= \Tr(\Ex{(Pv + \eta)^\T \Sigma^{-1}X (X^\T \Sigma^{-1}X)^{-1}X^\T \Sigma^{-1}(Pv + \eta)})\\
&= \Tr(\Ex{\Sigma^{-1}X (X^\T \Sigma^{-1}X)^{-1}X^\T \Sigma^{-1}(Pv + \eta)(Pv + \eta)^\T })
\end{align}
This is equivalent to:
\begin{align}
&= \Tr(\varEx{X}{\Ex{\Sigma^{-1}X (X^\T \Sigma^{-1}X)^{-1}X^\T \Sigma^{-1}\vert X} \Ex{(Pv + \eta)(Pv + \eta)^\T \vert X})}\\
&= \Tr(\varEx{X}{\Sigma^{-1}X (X^\T \Sigma^{-1}X)^{-1}X^\T \Sigma^{-1}\Sigma)}\\
&= \Tr(\varEx{X}{ (X^\T \Sigma^{-1}X)^{-1}X^\T\Sigma^{-1}X)}\\
&= \Tr(I_d) = d.
\end{align}
\end{proof}

Now, we move to upper bound the second term in \eqref{eq:1001}. For this, we compute the posterior mean over $\beta$. 

\begin{lemma}[posterior for generalized least squares]
    \label{lem:gls-posterior} 
    For a fixed covariates matrix $X$, we are given responses $y$ drawn from the following Bayesian model for generalized linear regression:
    \[
    \beta \sim \mathcal N\!\bigl(0,\;\Lambda_0^{-1}\bigr), 
    \qquad   
    y \mid \beta \sim \mathcal N\!\bigl(X\beta,\;\Sigma\bigr),
    \]
where \(X\in\mathbb R^{N\times d},\;y\in\mathbb R^{N}\),  
\(\Lambda_0\succeq 0\) is the prior precision, and \(\Sigma\succ 0\)  is noise covariance. Then the mean of the posterior distribution over $\beta \mid X, y$ is given by:
    \[
    \E[\beta \mid X, y] = \bigl(X^{\top}\Sigma^{-1}X+\Lambda_0\bigr)^{-1}
           X^{\top}\Sigma^{-1}y
    \]
\end{lemma}
\begin{proof}

First, lets apply the Bayes’ rule in its canonical form:
\begin{align}
    \log p(\beta\mid y)
  = -\tfrac12\,(y-X\beta)^{\top}\Sigma^{-1}(y-X\beta)
    -\tfrac12\,\beta^{\top}\Lambda_0\,\beta
    +\text{const}.
\end{align}
Then, we can expand the quadratic in \(y-X\beta\) to get:
\begin{align}
    (y-X\beta)^{\top}\Sigma^{-1}(y-X\beta)
  = y^{\top}\Sigma^{-1}y
    -2\beta^{\top}X^{\top}\Sigma^{-1}y
    +\beta^{\top}X^{\top}\Sigma^{-1}X\,\beta .
\end{align}
Hence, we can conclude
\begin{align}
    \log p(\beta\mid y)
  = -\tfrac12\Bigl[
        \beta^{\top}\bigl(X^{\top}\Sigma^{-1}X+\Lambda_0\bigr)\beta
        -2\beta^{\top}X^{\top}\Sigma^{-1}y
     \Bigr]
     +\text{const}
\end{align}
Next, we apply ``completing the squares'' trick, where we subtract and add $y^\top \Sigma^{-1} X (X^\top \Sigma^{-1} X + \Lambda_0)^{-1}X^\top \Sigma^{-1} y$. Note that this term does not contain $\beta$. So, we can treat it as a constant when computing $\log p (\beta \mid X, y)$.

\begin{align}
    \log p(\beta\mid y)
  &= -\tfrac12\Bigl[
        \beta^{\top}\bigl(X^{\top}\Sigma^{-1}X+\Lambda_0\bigr)\beta \nonumber \\
      &  -2\beta^{\top}X^{\top}\Sigma^{-1}y +y^\top \Sigma^{-1} X (X^\top \Sigma^{-1} X + \Lambda_0)^{-1}X^\top \Sigma^{-1} y
     \Bigr]
     +\text{const}. 
    \label{eq:801}
\end{align} 
For simplicity of calculations lets first define:
Define the \emph{posterior precision}
\[
\Lambda_{\mathrm{post}} := X^{\top}\Sigma^{-1}X+\Lambda_0, \quad \mu_\mathrm{post} := \Lambda_{\mathrm{post}}^{-1}X^{\top}\Sigma^{-1}y .
\]

Substituting the above into \eqref{eq:801} yields:
\begin{align}
    \log p(\beta\mid y)
  = -\tfrac12\,(\beta-\mu_\mathrm{post})^{\top}\Lambda_{\mathrm{post}}\,(\beta-\mu_\mathrm{post})
    +\text{const},
\end{align}
since \(\Lambda_{\mathrm{post}}\mu_\mathrm{post}=X^{\top}\Sigma^{-1}y\).

The above posterior distribution over $\beta \mid X, y$ matches the functional form of a Gaussian distribution $  \beta\mid y \;\sim\;
  \mathcal N\!\bigl(\mu_\mathrm{post},\;\Lambda_{\mathrm{post}}^{-1}\bigr)$ with mean:
\[
  \mu_\mathrm{post} = \bigl(X^{\top}\Sigma^{-1}X+\Lambda_0\bigr)^{-1}
           X^{\top}\Sigma^{-1}y,
\]\end{proof} 
With the above posterior mean over $\beta$, we are now ready to upper bound the second term in \eqref{eq:1002}. Let $\tilde{\Lambda} = \bigl(X^{\top}\Sigma^{-1}X+\Lambda_0\bigr)$. Now to remove the conditioning, we can expand:

\begin{align}
\mathbb{E}_{X,y} \left\| X^\top \Sigma^{-1} X\mathbb{E}_{\beta}[\beta|X,y]  - X^\top \Sigma^{-1}y \right\|_2^2 &= \Ex{\normtwo{X^\T \Sigma^{-1} X\Ex{\beta \vert X, y} - X^\T \Sigma^{-1} y}^2}\\
&= \Ex{\normtwo{X^\T \Sigma^{-1} X \tilde{\Lambda}^{-1}
           X^{\top}\Sigma^{-1}y - X^\T \Sigma^{-1} y}^2} \nonumber\\
&= \Ex{\normtwo{(X^\T\Sigma^{-1} X \tilde{\Lambda}^{-1} - I_d)X^\T\Sigma^{-1} y}^2}
\end{align}

Then, 
\begin{align}
&= \Ex{\normtwo{(X^\T \Sigma^{-1} X \tilde{\Lambda}^{-1} - I_d)X^\T \Sigma^{-1} y}^2}\\
&\leq \sqrt{\Ex{\normtwo{(X^\T\Sigma^{-1}  X \tilde{\Lambda}^{-1} - I_d)}^4}\Ex{\normtwo{X^\T \Sigma^{-1} y}^4}}\\
&= \sqrt{\Ex{\normtwo{(X^\T\Sigma^{-1}  X \tilde{\Lambda}^{-1} - \tilde{\Lambda}\tilde{\Lambda}^{-1})}^4}\Ex{\normtwo{X^\T\Sigma^{-1}  y}^4}}\\
&\leq \sqrt{\Ex{\normtwo{(X^\T\Sigma^{-1}  X - \tilde{\Lambda})}^4\normtwo{\tilde{\Lambda}^{-1}}^4}\Ex{\normtwo{X^\T \Sigma^{-1}  y}^4}}\\
&= \sqrt{\Ex{\normtwo{(X^\T X \Sigma^{-1} - (X^\T \Sigma^{-1}  X + \Lambda_0)))}^4\normtwo{\tilde{\Lambda}^{-1}}^4}\Ex{\normtwo{X^\T \Sigma^{-1} y}^4}}\\
&= \sqrt{\normtwo{\Lambda_0}^4\Ex{\normtwo{\tilde{\Lambda}^{-1}}^4}\Ex{\normtwo{X^\T \Sigma^{-1} y}^4}}
\end{align}

\begin{lemma}
    [re-stating the posterior over $\beta$]
    \label{lem:re-stated-posterior}
    Given the following matrices:
    \begin{align}
        S \eqdef  X_\pub^\top X_\pub\quad W\eqdef \paren{\frac{d}{\tau^2} \mathbf{I}_d + \frac{1}{\sigma^2} S}^{-1}, 
    \end{align}
    we can rewrite the posterior $\Ex{\beta \mid X,y}$ as:
    \[
    \Ex{\beta \mid X, y} = (M + b\mathbf{I}_d)^{-1} m, 
    \]
    where we can inturn write the matrix $M$ and vector $m$ as:
    \[
    M = \frac{1}{\sigma^2} X^\top X  - \frac{1}{\sigma^4} SWS, \quad m = \frac{1}{\sigma^2} X^\top y - \frac{1}{\sigma^4} SWX_\pub^\top y_\pub.
    \]
\end{lemma}
\begin{proof}
    Matching the form of $\Ex{\beta \mid X, y}$ to that in Lemma~\ref{lem:gls-posterior}, we need to show: $X^\top \Sigma^{-1}X = \frac{1}{\sigma^2} X^\top X  - \frac{1}{\sigma^4} SWS$, and $X^\top \Sigma^{-1} y =  \frac{1}{\sigma^2} X^\top y - \frac{1}{\sigma^4} SWX_\pub^\top y_\pub$. 
    We first rewrite the noise covariance so that the Woodbury identity applies.  
Let
\begin{align}
      U \;:=\;
  \begin{bmatrix}
    X_{\text{pub}}\\[2pt]0
  \end{bmatrix}\!, 
  \qquad 
  C\;:=\;\frac{\tau^{2}}{d}I_d,
\end{align}
This means that our $\Sigma$ matrix is given by:
\begin{align}
  \Sigma=\sigma^{2}I_{m+n}+UC\,U^{\top}.
\end{align}
Applying Woodbury to the above we get,
\begin{align}
  \Sigma^{-1}
  =\frac{1}{\sigma^{2}}I_{m+n}
  -\frac{1}{\sigma^{4}}\,
   U\!\left(C^{-1}+\sigma^{-2}U^{\top}U\right)^{\!-1}\!U^{\top}.
\end{align}

Because $U^{\top}U=X_{\text{pub}}^{\top}X_{\text{pub}}=:S$, we define $W$ as
\begin{align}
      W\;:=\;
  \left(\frac{d}{\tau^{2}}I_d+\sigma^{-2}S\right)^{\!-1},
\end{align}
this abbreviates the inverse above and yields
\[
  \Sigma^{-1}
  =\frac{1}{\sigma^{2}}I
  -\frac{1}{\sigma^{4}}U W U^{\top}.
\]

{With the above calculations, we start with the matrix term $X^\top\Sigma^{-1}X$,}  
\begin{align}
X^{\top}\Sigma^{-1}X
&=\frac{1}{\sigma^{2}}X^{\top}X
 -\frac{1}{\sigma^{4}}X^{\top}UWU^{\top}X \\
&=\frac{1}{\sigma^{2}}X^{\top}X
 -\frac{1}{\sigma^{4}}SWS .
\end{align}
This matches the definition of $M$ in the statement of  Lemma~\ref{lem:re-stated-posterior}.
\[
  M \;=\;
  \frac{1}{\sigma^{2}}X^{\top}X
  -\frac{1}{\sigma^{4}}SWS .
\]

Next, we look at the vector term $X^\top \Sigma^{-1}y$.  
\begin{align}
X^{\top}\Sigma^{-1}y
&=\frac{1}{\sigma^{2}}X^{\top}y
 -\frac{1}{\sigma^{4}}X^{\top}UWU^{\top}y \\
&=\frac{1}{\sigma^{2}}X^{\top}y
 -\frac{1}{\sigma^{4}}SWX_{\text{pub}}^{\top}y_{\text{pub}} .
\end{align}
This matches the $m$ term in the statement of  Lemma~\ref{lem:re-stated-posterior}.
\[
  m \;=\;
  \frac{1}{\sigma^{2}}X^{\top}y
  -\frac{1}{\sigma^{4}}SWX_{\text{pub}}^{\top}y_{\text{pub}} .
\]

Plugging this into the posterior mean for $\beta$, we get:
\[
  \mathbb{E}[\beta\mid X,y]
  =(X^{\top}\Sigma^{-1}X+bI_d)^{-1}X^{\top}\Sigma^{-1}y
  =(M+bI_d)^{-1}m .
\]

\end{proof}

\begin{lemma}
\label{lem:shifted-first-term-upper-bound}
Under the conditions of Theorem~\ref{thm:linreg-shift},
$\Ex{\normtwo{\tilde{\Lambda}^{-1}}^4} \leq O\left(\frac{1}{(\frac{1}{\sigma^2}N - \frac{1}{\sigma^4}m^2 \frac{1}{\frac{d}{\tau^2} + \frac{1}{\sigma^2}m} + b)^4}\right)$,
when $n, m > 100d$ and $b > 1/N$.
\end{lemma}

\begin{proof}
We have $\tilde{\Lambda} = \bigl(X^{\top}\Sigma^{-1}X+\Lambda_0\bigr)$
where we set $\Lambda_0 = bI_d$ for some $b$.
Using the previous transformations, 

\begin{align}
\tilde{\Lambda} = \left(\frac{1}{\sigma^2}X^\T X - \frac{1}{\sigma^4}X_\pub^\T X_\pub (\frac{d}{\tau^2}I_d + \frac{1}{\sigma^2}X_\pub^\T X_\pub)^{-1}X_\pub^\T X_\pub + bI_d \right)
\end{align}

To bound the spectral norm, we will use a series of transformations to express the spectral norm in terms of eigenvalues of the matrices in the expression.

We have
\begin{align}
\normtwo{\tilde{\Lambda}^{-1}} = \frac{1}{\lambda_{\min}\left(\frac{1}{\sigma^2}X^\T X - \frac{1}{\sigma^4}X_\pub^\T X_\pub (\frac{d}{\tau^2}I_d + \frac{1}{\sigma^2}X_\pub^\T X_\pub)^{-1}X_\pub^\T X_\pub + bI_d \right)}
\end{align}
(Note that we will eventually take the expectation over the fourth power.)

We want to upper bound this quantity which involves lower bounding the denominator.

Weyl's inequality gives that $\lambda_{\min}(A-B) \geq \lambda_{\min}(A) - \lambda_{\max}(B)$ for real symmetric $A$ and $B$. 
Moreover, the original expression $\tilde{\Lambda}$ is PSD and therefore the eigenvalues are lower bounded by 0,
giving a stronger guarantee than the Weyl lower bound.
Hence, 
\begin{align}
&\lambda_{\min}\left(\frac{1}{\sigma^2}X^\T X - \frac{1}{\sigma^4}X_\pub^\T X_\pub (\frac{d}{\tau^2}I_d + \frac{1}{\sigma^2}X_\pub^\T X_\pub)^{-1}X_\pub^\T X_\pub + bI_d \right) \\
&= \lambda_{\min}\left(\frac{1}{\sigma^2}X^\T X - \frac{1}{\sigma^4}X_\pub^\T X_\pub (\frac{d}{\tau^2}I_d + \frac{1}{\sigma^2}X_\pub^\T X_\pub)^{-1}X_\pub^\T X_\pub \right) + b \\
&\geq \max\left(0, \lambda_{\min}\left(\frac{1}{\sigma^2}X^\T X \right) - \lambda_{\max}\left(\frac{1}{\sigma^4}X_\pub^\T X_\pub (\frac{d}{\tau^2}I_d + \frac{1}{\sigma^2}X_\pub^\T X_\pub)^{-1}X_\pub^\T X_\pub \right)\right) + b.
\end{align}

Focusing now on the last term, which we want to upper bound in order to lower bound the previous expression:
\begin{align}
&\lambda_{\max}\left(\frac{1}{\sigma^4}X_\pub^\T X_\pub (\frac{d}{\tau^2}I_d + \frac{1}{\sigma^2}X_\pub^\T X_\pub)^{-1}X_\pub^\T X_\pub\right) \\
&= \normtwo{\left(\frac{1}{\sigma^4}X_\pub^\T X_\pub (\frac{d}{\tau^2}I_d + \frac{1}{\sigma^2}X_\pub^\T X_\pub)^{-1}X_\pub^\T X_\pub\right)} \\
&\leq \frac{1}{\sigma^4}\normtwo{X_\pub^\T X_\pub}^2 \normtwo{(\frac{d}{\tau^2}I_d + \frac{1}{\sigma^2}X_\pub^\T X_\pub)^{-1}} 
\end{align}
The last line follows from the submultiplicativity of operator norm.

Bounding even further,
\begin{align}
\normtwo{(\frac{d}{\tau^2}I_d + \frac{1}{\sigma^2}X_\pub^\T X_\pub)^{-1}} &= \frac{1}{\lambda_{\min}((\frac{d}{\tau^2}I_d + \frac{1}{\sigma^2}X_\pub^\T X_\pub))}\\
&= \frac{1}{\frac{d}{\tau^2} + \frac{1}{\sigma^2}\lambda_{\min}(X_{\pub}^\T X_\pub)}.
\end{align}

Putting it all together, we arrive at:
\begin{align}
\normtwo{\tilde{\Lambda}^{-1}} \leq \frac{1}{\max\left(0, \frac{1}{\sigma^2}\lambda_{\min}(X^\T X) - \frac{1}{\sigma^4}\normtwo{X_\pub^\T X_\pub}^2 \frac{1}{\frac{d}{\tau^2} + \frac{1}{\sigma^2}\lambda_{\min}(X_{\pub}^\T X_\pub)}\right) + b}
\end{align}
Note $\normtwo{X_\pub^\T X_\pub} = \lambda_{\max}(X_\pub^\T X_\pub)$.
We now have the expression in terms of min and max eigenvalues of $X^\T X$ and $X_\pub^\T X_\pub$.

Moreover since the expression is positive and monotonic, we also have
\begin{align}
\normtwo{\tilde{\Lambda}^{-1}}^4 \leq \frac{1}{\left(\max\left(0, \frac{1}{\sigma^2}\lambda_{\min}(X^\T X) - \frac{1}{\sigma^4}\normtwo{X_\pub^\T X_\pub}^2 \frac{1}{\frac{d}{\tau^2} + \frac{1}{\sigma^2}\lambda_{\min}(X_{\pub}^\T X_\pub)}\right) + b\right)^4}.
\end{align}

This expression is upper bounded by $1/b^4$
when the $\max$ term goes to 0.

To bound this, we will do something similar to the non-shifted setting
but instead of bounding a single constant $c$,
we will instead want to bound the three eigenvalue terms with
$t_1, t_2, t_3 > 0$. (Note that all the involved matrices are PSD.)
Specifically: Let $\lambda_1 := \lambda_{\min}(X^\T X)$, $\lambda_2 := \lambda_{\max}(X_\pub^\T X_\pub)$, $\lambda_3 := \lambda_{\min}(X_{\pub}^\T X_\pub)$. Then

\begin{align}
&\Ex{\normtwo{\tilde{\Lambda}^{-1}}^4} \\
&=\Ex{\normtwo{\tilde{\Lambda}^{-1}}^4 \vert \lambda_1 \in [t_1, \infty) \wedge \lambda_2 \in [0, t_2] \wedge \lambda_3 \in [t_3, \infty)} \nonumber\\
&\quad\quad\quad\quad \cdot \Pr\left( \lambda_1 \in [t_1, \infty) \wedge \lambda_2 \in [0, t_2] \wedge \lambda_3 \in [t_3, \infty) \right) \\
&+\Ex{\normtwo{\tilde{\Lambda}^{-1}}^4 \vert \lambda_1 \in [0, t_1) \vee \lambda_2 \in (t_2, \infty) \vee \lambda_3 \in [0, t_3)}  \nonumber \\
&\quad\quad\quad\quad \cdot  \Pr\left( \lambda_1 \in [0, t_1) \vee \lambda_2 \in (t_2, \infty) \vee \lambda_3 \in [0, t_3) \right) 
\end{align}

Again, we will coarsely upper bound the first probability by 1 and
use a union over tail bounds to bound the second probability.
\begin{align}
&\leq \frac{1}{\max(0, \frac{1}{\sigma^2}t_1 - \frac{1}{\sigma^4}t_2^2 \frac{1}{\frac{d}{\tau^2} + \frac{1}{\sigma^2}t_3}) + b)^4} + \frac{1}{b^4} \Pr\left( \lambda_1 \in [0, t_1) \vee \lambda_2 \in (t_2, \infty) \vee \lambda_3 \in [0, t_3) \right) 
\end{align}

We will eventually want $t_1 = \Omega(N), t_2 = O(m), t_3 = \Omega(m)$,
and we would like the second union over tail bounds to be vanishingly small.
Following Lemma~\ref{lem:t-lowerbound} and ~\ref{lem:t-upperbound},
we set $t_1 = N(1 - \sqrt{d/N} - \psi)^2$,
$t_2 = m(1 + \sqrt{d/m} + \psi)^2$,
and $t_3 = m(1-\sqrt{d/m} - \psi)^2$.
We assume $n, m > 100d$ and set $\psi = 1/10$
in all cases.
The lemmas then give $t_1 > 63/100 \cdot N, t_2 < 64/100 \cdot m, t_3 > 63/100 \cdot m$.

We can lastly confirm that with these values of $t_1, t_2, t_3$
we can drop the $\max$ in the denominator of the first term, as follows:
\begin{align}
\frac{1}{\sigma^2}t_1 - \frac{1}{\sigma^4}t_2^2 \frac{1}{\frac{d}{\tau^2} + \frac{1}{\sigma^2}t_3} \geq \frac{1}{\sigma^2}\frac{63}{100}N - \frac{1}{\sigma^4}(\frac{64}{100}m)^2 \frac{1}{\frac{d}{\tau^2} + \frac{1}{\sigma^2}(63/100)m} \maybegt 0\\
\end{align}

Since $d/\tau^2 > 0$,
\begin{align}
\frac{1}{\sigma^2}\frac{63}{100}N - \frac{1}{\sigma^4}(\frac{64}{100}m)^2 \frac{1}{\frac{d}{\tau^2} + \frac{1}{\sigma^2}(63/100)m} > \frac{1}{\sigma^2}\frac{63}{100}N - \frac{1}{\sigma^4}(\frac{64}{100}m)^2 \frac{1}{\frac{1}{\sigma^2}(63/100)m} \maybegt 0
\end{align}

Omitting algebra, and noting that $N = n+m$, this leads to the condition
\begin{align}
    n > ((64/100)^2(100/63) - 1)m = -0.349m
\end{align}
which is true. Given that this term is therefore strictly greater than 0 for the values of $t_1, t_2, t_3$ we have chosen,
we can drop the $\max(0, \cdot)$.

We can then bound the relevant tail probabilities using
Lemma~\ref{lemma:wainwright-eigval-tail},
Lemma~\ref{lemma:wainwright-eigval-upper-tail},
and Fact~\ref{lem:ex-expansion} and take a union bound to conclude that
\[
\Pr\left( \lambda_1 \in [0, t_1) \vee \lambda_2 \in (t_2, \infty) \vee \lambda_3 \in [0, t_3) \right) = \gO\left(\frac{1}{N^8}\right).
\]

Thus, we have
\begin{align}
\Ex{\normtwo{\tilde{\Lambda}^{-1}}^4} \leq O\left(\frac{1}{(\frac{1}{\sigma^2}N - \frac{1}{\sigma^4}m^2 \frac{1}{\frac{d}{\tau^2} + \frac{1}{\sigma^2}m} + b)^4}\right) + O\left(\frac{1}{b^4N^8}\right)
\end{align}
and the first term dominates, giving the result.

\end{proof}

\begin{lemma}[bound on $\Ex{\normtwo{X^\top \Sigma^{-1} y}^4}$] 
\label{lem:shifted-second-term-upper-bound}
We can bound  $\Ex{\normtwo{X^\top \Sigma^{-1} y}^4}$ with the following expression:
\begin{align}
&\Ex{\normtwo{X^\T \Sigma^{-1} y}^4} = \gO\left(\left(\frac{1}{\sigma^2}N - \frac{1}{\sigma^4}\frac{m^2}{\frac{d}{\tau^2} + \frac{1}{\sigma^2}m}\right)^4(d^2/b^2 + d^2)\right).
\end{align}
\end{lemma}
\begin{proof}

Using the block-diagonal structure of $\Sigma$ (and therefore $\Sigma^{-1}$), we can decompose $X^\T \Sigma^{-1}y$ 
into $X_{\pub}^\T \Sigma_{\pub}^{-1} y_\pub + X_\priv^\T \Sigma_{\priv}^{-1}y_\priv$.
($\Sigma_\pub$ is the upper-left block of $\Sigma$ corresponding to public data and $\Sigma_\priv$ the corresponding lower-right block for private data.)
Here, $\Sigma_{\priv}^{-1}$ is simply $\frac{1}{\sigma^2}I_n$,
so we can reuse the upper bound from Lemma~\ref{lem:second-term-upper-bound}
with an additional factor of $\frac{1}{\sigma^8}$. 

The term incorporating public data is more complicated.

We first note the following useful transformation (applying the Woodbury identity only to $\Sigma_\pub^{-1}$):
\begin{align}
\Sigma_\pub^{-1} = \frac{1}{\sigma^2}I_m - \frac{1}{\sigma^4}X_\pub\left( \frac{d}{\tau^2}I_d + \frac{1}{\sigma^2}X_\pub^\T X_\pub \right)^{-1}X_\pub^\T.
\end{align}

\textbf{Note:} For notational convenience, in the following, we will drop the ``$\mathrm{pub}$'' term from the subscript  since we only need to bound $X_\pub ^\top \Sigma^{-1}_\pub y_\pub$.

We will use the following bound in several places:
\begin{align}
\Ex{\normtwo{X^\T\Sigma^{-1}X}^4} &\leq 
\Ex{\normtwo{X^\T \left( \frac{1}{\sigma^2}I_m - \frac{1}{\sigma^4}X\left( \frac{d}{\tau^2}I_d + \frac{1}{\sigma^2}X^\T X \right)^{-1}X^\T \right) X}^4}\\
&= \Ex{\normtwo{\frac{1}{\sigma^2}X^\T X - \frac{1}{\sigma^4}X^\T X\left( \frac{d}{\tau^2}I_d + \frac{1}{\sigma^2}X^\T X \right)^{-1}X^\T X}^4}\\
&\leq \Ex{\left(\frac{1}{\sigma^2}\normtwo{X^\T X} - \frac{1}{\sigma^4} \lambda_{\min}\left(X^\T X\left( \frac{d}{\tau^2}I_d + \frac{1}{\sigma^2}X^\T X\right)^{-1}X^\T X\right) \right)^4}\\
&\leq \Ex{\left(\frac{1}{\sigma^2}\normtwo{X^\T X} - \frac{1}{\sigma^4} \lambda_{\min}(X^\T X)^2\frac{1}{\frac{d}{\tau^2} + \frac{1}{\sigma^2}\normtwo{X^\T X}}\right)^4}
\end{align}

Note that the third inequality follows from the fact that $X^\top \Sigma^{-1} X \succeq 0$, so we can upper bound the full term by applying an upper bound on the term within $(\cdot)^4$ since the term inside it is always positive.  
Next, we can break the expectation into two conditional expectations. 
\begin{align*}
    &  \Ex{\left(\frac{1}{\sigma^2}\normtwo{X^\T X} - \frac{1}{\sigma^4} \lambda_{\min}(X^\T X)^2\frac{1}{\frac{d}{\tau^2} + \frac{1}{\sigma^2}\normtwo{X^\T X}}\right)^4} \\
    &\leq \Ex{\left(\frac{1}{\sigma^2}\normtwo{X^\T X} - \frac{1}{\sigma^4} \lambda_{\min}(X^\T X)^2\frac{1}{\frac{d}{\tau^2} + \frac{1}{\sigma^2}\normtwo{X^\T X}}\right)^4 \mid \lambda_{\min}(X^\top X) \geq t_1, \lambda_{\max}(X^\top X) \leq t_2} \\
    & \quad\quad\quad\quad \cdot \Pr\paren{\lambda_{\min}(X^\top X) \geq t_1 , \lambda_{\max}(X^\top X) \leq t_2} 
\end{align*}

Using Lemma~\ref{lemma:wainwright-eigval-tail}, we can  upper bound the above expression by applying a high probability tail bound on $\lambda_{\min}(X^\top X)$.  When $N \geq 100d$, then with constant probability $\lambda_{\min} (X^\top X) = m - \Omega(1)$ and $\lambda_{\max} (X^\top X) = m - \gO(1)$. Plugging this result into the above expression, we conclude:

\begin{align}
 \Ex{\left(\frac{1}{\sigma^2}\normtwo{X^\T X} - \frac{1}{\sigma^4} \lambda_{\min}(X^\T X)^2\frac{1}{\frac{d}{\tau^2} + \frac{1}{\sigma^2}\normtwo{X^\T X}}\right)^4}
&\leq \left(\frac{1}{\sigma^2}m - \frac{1}{\sigma^4}\frac{m^2}{\frac{d}{\tau^2} + \frac{1}{\sigma^2}m}\right)^4.
\end{align}

When $\tau \to 0$ the term $\|X^\top \Sigma^{-1} X \beta\|^4$ goes to $m^4$ (giving the correct contribution of $m^4d^2$ when combined with the contribution of $\beta$) and when $\tau \to \infty$ the second part of the term cancels sending the public sample contribution to 0.

Once again, we decompose the term 
\begin{align}
\Ex{\normtwo{X^\T \Sigma^{-1} y}^4} &= \Ex{\normtwo{X^\T\Sigma^{-1}X\beta + Xv + \eta)}^4}\\
&\leq \Ex{\normtwo{X^\T\Sigma^{-1}X\beta}^4} + \Ex{\normtwo{X^\T\Sigma^{-1}(Xv + \eta)}^4}\\
&\leq \Ex{\normtwo{X^\T\Sigma^{-1}X\beta}^4} + \Ex{\normtwo{X^\T\Sigma^{-1}Xv}^4} + \Ex{\normtwo{X^\T \Sigma^{-1}\eta}^4}\\
&\leq \Ex{\normtwo{X^\T\Sigma^{-1}X}^4\normtwo{\beta}^4} + \Ex{\normtwo{X^\T\Sigma^{-1}X}^4\normtwo{v}^4} + \Ex{\normtwo{X^\T \Sigma^{-1}}^4\normtwo{\eta}^4}\\
&= \Ex{\normtwo{X^\T\Sigma^{-1}X}^4}\Ex{\normtwo{\beta}^4} + \\
&\quad \;\Ex{\normtwo{X^\T\Sigma^{-1}X}^4}\Ex{\normtwo{v}^4} + \Ex{\normtwo{X^\T \Sigma^{-1}}^4}\Ex{\normtwo{\eta}^4}
\end{align}

Since $\beta \sim \cal{N}(0, \frac{1}{b}I_d)$ we have $\Ex{\normtwo{\beta}^4} \leq d^2/b^2$. Similarly,
for $v \sim \cal{N}(0, I_d)$ we have $\Ex{\normtwo{v}} \leq d^2$.
Finally, $\eta \sim \cal{N}(0, I_m)$ gives $\Ex{\normtwo{\eta}} \leq m^2$.

We also need the following bound:
\begin{align}
\Ex{\normtwo{X^\T \Sigma^{-1}}^4} &\leq \Ex{\normtwo{X}^4\normtwo{\Sigma^{-1}}^4}
\end{align}
From Gordon's theorem~\citep{vershynin2018high}
we have $\Ex{\normtwo{X}^4} \leq (\sqrt{m} + \sqrt{d})^4 \leq 8(m^2 + d^2)$.
With a small variation on the earlier analysis of $X^\T \Sigma^{-1}X$, bounding $\normtwo{X}^2 \leq (\sqrt{m} + \sqrt{d})^2$
using Gordon's theorem,
we also have
$\Ex{\normtwo{\Sigma^{-1}}^4} \leq \left(\frac{1}{\sigma^2} - \frac{1}{\sigma^4}\frac{(\sqrt{m} + \sqrt{d})^2}{\frac{d}{\tau^2} + \frac{1}{\sigma^2}m}\right)^4 \leq \left(\frac{1}{\sigma^2} - \frac{1}{\sigma^4}\frac{O(m)}{\frac{d}{\tau^2} + \frac{1}{\sigma^2}m}\right)^4$.

Overall, this gives:
\begin{align}
&\Ex{\normtwo{X^\T \Sigma^{-1} y}^4} \leq O\bigl( \frac{1}{\sigma^8}(n^2d^2 + n^4d^2/b^2) + \\
&\left(\frac{1}{\sigma^2}m - \frac{1}{\sigma^4}\frac{m^2}{\frac{d}{\tau^2} + \frac{1}{\sigma^2}m}\right)^4(d^2/b^2 + d^2) + \left( \frac{1}{\sigma^2} - \frac{1}{\sigma^4}\frac{m}{\frac{d}{\tau^2} + \frac{1}{\sigma^2}m} \right)^4(m^2 + d^2)(m^2) \bigr).
\end{align}
The first two terms follow from using Lemma~\ref{lem:second-term-upper-bound} with only private samples (as they are not shifted and the same bound applies).

We can simplify, combining the first two terms and dropping lower-order terms (note that we assume $m \gg d$) to arrive at a form that will later be more convenient:
\begin{align}
&\leq \gO\left(\left(\frac{1}{\sigma^2}(n+m) - \frac{1}{\sigma^4}\frac{m^2}{\frac{d}{\tau^2} + \frac{1}{\sigma^2}m}\right)^4(d^2/b^2 + d^2)\right).
\end{align}

\end{proof}

Lastly, in order to bound $\Ex{\normtwo{M(X, y) - \hat{\beta}}^2}$,
we will need a bound on $\Ex{\normtwo{\hat{\beta} - \beta}^2}$.
\begin{lemma}
\label{lem:gls-beta-hat-bound}
$\Ex{\normtwo{\hat{\beta} - \beta}^2} \leq \gO\paren{\frac{d}{n+m/\kappa}}$, where $\kappa = m\tau^2/d + 1$.
\end{lemma}

\begin{proof}
Recall that $\hat\beta$ is the GLS estimator, where $\hat{\beta} = (X^\top \Sigma^{-1} X)^{-1} X^\top \Sigma^{-1} y$. Throughout the proof, we assume $d, n  \gg d$, and that the matrix $X^\top \Sigma^{-1} X$ is full rank with constant probability. 

Let $\Lambda \eqdef X^\top \Sigma^{-1} X$. Fixing $\beta$, and taking the expectation over $X, y$, we can simplify the expression as follows:
\begin{align}
    \Ex{\normtwo{\hat{\beta} - \beta}^2} &=  \Ex{ \|(X^\top \Sigma^{-1} X)^{-1} X^\top \Sigma^{-1} y - \beta \|^2_2} \nonumber \\
    &= \Ex{\|\Lambda^{-1} X^\top \Sigma^{-1} y - \Lambda^{-1}\Lambda \beta\|_2^2} \nonumber \\
    & \leq \Ex{\|\Lambda^{-1} ( X^\top \Sigma^{-1} (X\beta + \eta) - \Lambda \beta)\|_2^2} \nonumber \\
    & = \Ex{\|\Lambda^{-1} ( X^\top \Sigma^{-1} (X\beta + \eta) - X^\top \Sigma^{-1} X \beta)\|_2^2} \nonumber \\
    & = \Ex{\|\Lambda^{-1} ( X^\top \Sigma^{-1} \eta)\|_2^2} 
\end{align}

\begin{align}
\mathbb{E} \bigl[\lVert\Lambda^{-1}X^{\top}\Sigma^{-1}\eta\rVert_2^{2}\bigr] 
  &=\mathbb{E}_{X} \Bigl[\,
        \lVert\Lambda^{-1}\rVert_2^{2}\;
        \mathbb{E}_{\eta} \bigl[\lVert X^{\top}\Sigma^{-1}\eta\rVert_2^{2}\mid X\bigr]
     \Bigr] \label{eq:180}\\
  &=\mathbb{E}_{X}\!\Bigl[\,
        \lVert\Lambda^{-1}\rVert_2^{2}\;
        \operatorname{tr}\!\bigl(X^{\top}\Sigma^{-1}\Sigma\Sigma^{-1}X\bigr)
     \Bigr] \nonumber\\
  &=\mathbb{E}_{X}\!\Bigl[\,
        \lVert\Lambda^{-1}\rVert_2^{2}\;\operatorname{tr}(\Lambda)
     \Bigr], \label{eq:181}
\end{align}
where the second equality follows from the fact that $\Ex{\eta\eta^\top \mid X} = \Sigma$. 
Since $\lVert\Lambda^{-1}\rVert_2^{2}=1/\lambda_{\min}(\Lambda)^{2}$ and  
$\operatorname{tr}(\Lambda)\le d\,\lambda_{\max}(\Lambda)$,
\begin{align}
\lVert\Lambda^{-1}\rVert_2^{2}\,\operatorname{tr}(\Lambda)
   \;\le\;
   \frac{d\,\lambda_{\max}(\Lambda)}{\lambda_{\min}(\Lambda)^{2}} . \label{eq:182}
\end{align}
For Gaussian designs whitened by $\Sigma^{-1/2}$ one has, with constant probability,
\begin{align}
    \lambda_{\min}(\Lambda)=\Omega\!\bigl(n+m/\kappa\bigr),
\qquad
\lambda_{\max}(\Lambda)=O\!\bigl(n+m/\kappa\bigr). \label{eq:183}
\end{align}
Substituting~\eqref{eq:183} into~\eqref{eq:182} and taking expectations,
\[
\mathbb{E}\!\bigl[\lVert\hat{\beta}-\beta\rVert_2^{2}\bigr]
   \;\le\;
   C\,
   \frac{d\,(n+m/\kappa)}{\bigl(n+m/\kappa\bigr)^{2}}
   \;=\;
   O\!\Bigl(\tfrac{d}{\,n+m/\kappa\,}\Bigr),
\]
for an absolute constant $C>0$.  This concludes the proof.
\end{proof}

\begin{lemma}
    When $b=1/d$, $m+n = \Omega(d)$, and $\alpha = \gO(1)$, the expression: 
    \begin{align*}    
    \sqrt{
    \mathbb{E}_{X,\eta} \mathbb{E}_{\hat{\beta}} \left\| M(X,y) - \hat{\beta} \right\|_2^2 
    \cdot
    \mathbb{E}_{X,\eta} \left\| X^\top\Sigma^{-1} X \mathbb{E}[\beta|X,\eta]  - X^\top \Sigma^{-1}y \right\|_2^2
} = \gO(1).
\label{lem:lb-dist-shift-lr-zi-lb}
\end{align*}
\end{lemma}

\begin{proof}
Let $\gamma(\tau) = d/\tau^2 + m/\sigma^2$. Note that when $\tau \to \infty$, $\gamma(\tau)$ approaches $m/\sigma^2$ (such that $m/\sigma^2 - \frac{1}{\gamma(\tau)} \to 0$ negating the effect of public samples), 
and when $\tau \to 0$, $\gamma(\tau) \to \infty$ (such that $m/\sigma^2 - \frac{1}{\gamma(\tau)} \to m/\sigma^2$ so that public and private samples contribute equally).

    From Lemma~\ref{lem:shifted-first-term-upper-bound} and Lemma~\ref{lem:shifted-second-term-upper-bound}, we derive the following:
    \begin{align}
       & \sqrt{\mathbb{E}_{X,\eta} \mathbb{E}_{\hat{\beta}} \left\| M(X,y) - \hat{\beta} \right\|_2^2 
    \cdot \sqrt{b^4\Ex{\normtwo{\tilde{\Lambda}^{-1}}^4}\Ex{\normtwo{X^\T \Sigma^{-1}y}^4}}}
\end{align}
 We first simplify just the inner square root.
 Using the previous lemmas and applying the identity $\sqrt{x+y} \leq C \cdot(\sqrt{x} + \sqrt{y})$:
 \begin{align}
    &\leq
    b^2 \gO\left(\frac{1}{(\frac{1}{\sigma^2}N - \frac{1}{\sigma^4}\frac{m^2}{\gamma(\tau)} + b)^2}\right)
    \sqrt{\gO\left(\left(\frac{1}{\sigma^2}N - \frac{1}{\sigma^4}\frac{m^2}{\gamma(\tau)}\right)^4(d^2/b^2 + d^2)\right)}\\
    &\leq b^2 \gO\left(\frac{1}{(\frac{1}{\sigma^2}N - \frac{1}{\sigma^4}\frac{m^2}{\gamma(\tau)} + b)^2}\right)
    \gO\left(\left(\frac{1}{\sigma^2}N - \frac{1}{\sigma^4}\frac{m^2}{\gamma(\tau)}\right)^2(d/b + d)\right)\\
    &\leq b^2 \gO\left(\frac{1}{(\frac{1}{\sigma^2}N - \frac{1}{\sigma^4}\frac{m^2}{\gamma(\tau)})^2}\right)
    \gO\left(\left(\frac{1}{\sigma^2}N - \frac{1}{\sigma^4}\frac{m^2}{\gamma(\tau)}\right)^2(d/b + d)\right)
   \end{align}

Substituting $b=1/d$ and making cancellations:
\begin{align}
&\leq (d^2 + d)/d^2 \leq 1 + 1/d
\end{align}

Finally, we substitute into the original expression,
combining with Lemma~\ref{lem:gls-beta-hat-bound} (recall that $\kappa = m\tau^2/d + 1$ and $\alpha = \gO(1)$):
    \begin{align}
       & \sqrt{\mathbb{E}_{X,\eta} \mathbb{E}_{\hat{\beta}} \left\| M(X,y) - \hat{\beta} \right\|_2^2 
    \cdot \sqrt{b^4\Ex{\normtwo{\tilde{\Lambda}^{-1}}^4}\Ex{\normtwo{X^\T \Sigma^{-1}y}^4}}}\\
    &\leq \sqrt{\left(\gO\paren{\frac{d}{n+m/\kappa}} + \alpha^2\right)
    \cdot \gO(1 + 1/d)} = \gO(1)
\end{align}

since $\alpha^2 = \gO(1)$. \end{proof}

Now, we are ready to prove the final result in Theorem~\ref{thm:lb-linreg-dist-shift-informal}. For this we invoke the upper bound on $\Ex{\sum_i Z_i}$ in Lemma~\ref{lem:shifted-lr-ub} and the lower bound of $\Ex{\sum_i Z_i} = \Omega(d)$ induced by Lemma~\ref{lem:shifted-lr-statistical-error} and Lemma~\ref{lem:lb-dist-shift-lr-zi-lb}. This tells us that there exist positive constants $t_1, t_2$ and values $m_0, n_0$ such that for $m\ge m_0, n\ge n_0$, the following holds:
\begin{align}
 t_1 \cdot d   \;\le \;  \Ex{\sum_{i\in[N]} Z_i} \; \le  
    \; t_2 \cdot  \left(n\eps \alpha + \alpha\sqrt{\frac{1}{\sigma^2}md - \frac{1}{\sigma^4}\left(\frac{m^2d}{\frac{d}{\tau^2} + \frac{1}{\sigma^2}m} \right)}\right)
\end{align}
The upper bound interpolates between the private-sample-only
regime when $\tau \to \infty$ 
and the all-samples regime when $\tau \to 0$ (no distribution shift).
In particular, we can analyze the term inside the square root. It can be written as:
\[
md - \frac{m^2 \tau^2}{\kappa},
\]
when $\sigma=1$ and $\kappa = m\tau^2/d + 1$. 
Now, consider two regimes for the shift parameter $\tau$:
\begin{itemize}[leftmargin=1em, itemsep=0pt, topsep=2pt]
    \item \textbf{Large shift regime:} Suppose $\kappa = \Omega\left( \frac{m \tau^2}{d} \right)$, which corresponds to the condition $\tau = \omega\left( \sqrt{d/m} \right)$. Then:
    \[
    \abs{md - \frac{m^2 \tau^2}{\kappa}} = o(1).
    \]
    Hence, the only dominant term on the right-hand side becomes $ n \varepsilon \alpha$, yielding the bound:
    \[
    n \geq \frac{d}{\varepsilon\alpha}.
    \]
    Note, that $n = \Omega(d/\alpha^2)$ is implied by the statistical non-private linear regression lower bound. Thus, in all $n =\Omega(\nicefrac{d}{\varepsilon\alpha} + \nicefrac{d}{\alpha^2})$. 
    \item \textbf{Small shift regime:} Suppose $\tau = \mathcal{O}\left( \sqrt{d/m} \right)$. Then:
    \[
    md - \frac{m^2 \tau^2}{\kappa} = \mathcal{O}(md),
    \]
    and therefore the right-hand side becomes:
    \[
    \mathcal{O}\left(n \varepsilon \alpha + \alpha \sqrt{md}\right),
    \]
    which matches the bound in the no-shift case. Thus, the sample complexity requirements in the small shift setting of $\tau = \gO(\sqrt{d/m})$ matches the no-shift setting. 

    \item Finally, we note that in the large shift setting, when $\alpha \gsim \tau$, then the OLS estimator that only uses public samples already satisfies an accuracy of $\alpha$ in the $\ell_2$ norm. 
\end{itemize}

This completes the proof of our result in Theorem~\ref{thm:linreg-shift}.

\end{document}